\documentclass{article}
\usepackage[final,nonatbib]{neurips_2024}
\usepackage[utf8]{inputenc} 
\usepackage[T1]{fontenc}    
\usepackage{hyperref}       
\hypersetup{hidelinks=true}
\usepackage{url}            
\usepackage{booktabs}       
\usepackage{nicefrac}       
\usepackage{microtype}      
\usepackage{xcolor}         
\usepackage{listings}%
\usepackage{colortbl}
\usepackage{makecell}
\usepackage{subfigure}
\usepackage{graphicx}%
\usepackage{multirow}%
\usepackage{amsmath,amssymb,amsfonts}%
\usepackage{amsthm}%
\usepackage{mathrsfs}%
\usepackage{textcomp}%
\usepackage{algorithm}%
\usepackage{algorithmicx}%
\usepackage{algpseudocode}%
\newcommand{\ie}{i.e.,\ }

\newcommand{\et}{\emph{et al.}\ }

\title{Towards Adaptive Meta-Gradient Adversarial Examples for Visual Tracking}

\author{
Wei-Long Tian$^1$, Peng Gao$^1$, Xiao Liu$^1$, Long Xu$^1$, Hamido Fujita$^{2,3,4,5}$,\\
\textbf{Hanan Aljuai}$^3$, \textbf{Mao-Li Wang}$^1$\\
$^1$School of Cyber Science and Engineering, Qufu Normal University, China\\
$^2$Malaysia-Japan International Institute of Technology, University Teknologi Malaysia, Malaysia\\
$^3$Computer Sciences Department, Princess Nourah bint Abdulrahman University, Saudi Arabia\\
$^4$Regional Research Center, Iwate Prefectural University, Japan\\
$^5$Faculty of Science, University of Hradec Kralove, Czech Republic
}

\begin{document}

\maketitle

\begin{abstract}
In recent years, visual tracking methods based on convolutional neural networks and Transformers have achieved remarkable performance and have been successfully applied in fields such as autonomous driving. However, the numerous security issues exposed by deep learning models have gradually affected the reliable application of visual tracking methods in real-world scenarios. Therefore, how to reveal the security vulnerabilities of existing visual trackers through effective adversarial attacks has become a critical problem that needs to be addressed. To this end, we propose an adaptive meta-gradient adversarial attack (AMGA) method for visual tracking. This method integrates multi-model ensemble and meta-learning strategies, combining momentum mechanisms and Gaussian smoothing, which can significantly enhance the transferability and attack effectiveness of adversarial examples. AMGA randomly selects models from a large model repository, constructs diverse tracking scenarios, and iteratively performs both white- and black-box adversarial attacks in each scenario, optimizing the gradient directions of each model. This paradigm minimizes the gap between white- and black-box adversarial attacks, thus achieving excellent attack performance in black-box scenarios. Extensive experimental results on large-scale datasets such as OTB2015, LaSOT, and GOT-10k demonstrate that AMGA significantly improves the attack performance, transferability, and deception of adversarial examples. Codes and data are available at \url{https://github.com/pgao-lab/AMGA}.
\end{abstract}

\section{Introduction}\label{sec:1}

Visual tracking is one of most fundamental and crucial research topics in computer vision\cite{r1}. It not only plays a key role in tasks such as scene understanding, intelligent surveillance, and autonomous driving. It also profoundly impacts the advancement of various fields, including human-computer interaction and medical image analysis. In recent years, deep learning models has flourished in computer vision applications, with the rapid development of convolutional neural networks (CNNs) significantly enhancing the performance of visual tracking to unprecedented levels\cite{r3,r4}. Deep learning-based visual tracking methods\cite{rar,r8,satin,r6,siamextr} automatically learn high-dimensional feature representations from images, enabling trackers to exhibit higher robustness and adaptability in complex scenarios, outperforming traditional methods. However, Goodfellow \et revealed that deep learning models are highly vulnerable when faced with adversarial examples, directly exposing the security and robustness issues of deep learning models in practical applications\cite{r2,r11}. Researchers have found that even imperceptible perturbations can severely degrade overall performance in visual tracking. When subjected to adversarial attacks, an ideal visual tracker should demonstrate high performance in security and robustness. Therefore, enhancing the effectiveness and transferability of adversarial examples to uncover the security vulnerabilities of existing visual trackers and provide directions for improving the security and robustness of visual tracking is an urgent research problem.

Traditional adversarial attack methods, such as the fast gradient sign method (FGSM) \cite{r2} and the iterative FGSM (I-FGSM) \cite{r11}, utilize gradient information of deep learning models through carefully designed perturbations, successfully generating adversarial examples practical in white-box scenarios. However, these methods mainly focus on object recognition and detection in static and discrete images. Generating compelling adversarial examples becomes significantly more challenging when this research perspective shifts to dynamic and continuous tasks. Visual tracking requires locating the target objects in a single frame and demands consistent detection across consecutive frames. Frame-by-frame optimization methods such as FGSM and I-FGSM typically fail to capture the temporal dependencies and dynamic adjustment capabilities of trackers. Consequently, the adversarial examples they generate may impact tracking performance in the short term but struggle to consistently disrupt tracking robustness. Moreover, the increasing complexity of visual trackers exacerbates the difficulty of generating adversarial examples. Different trackers may employ vastly different feature extraction, object representation, and detection strategies, requiring adversarial examples for stronger transferability. However, many existing attack methods rely on gradient information from a single model, leading to adversarial examples with poor transferability across different models. This limitation is particularly evident in black-box scenarios, where practical applications are constrained, rendering effective attacks nearly impossible.

\begin{figure}[t]
	\begin{center}
		\includegraphics[width=0.7\linewidth]{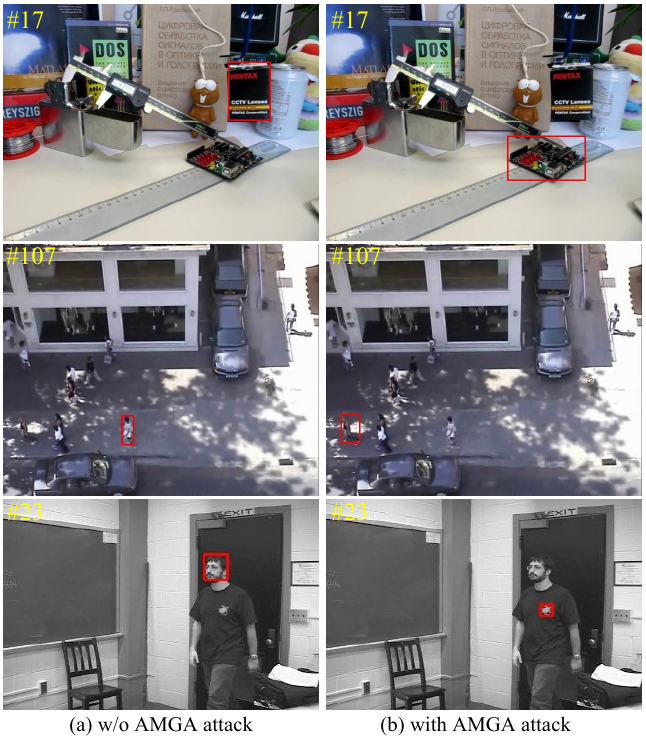}
	\end{center}
	\vspace{-0.8em}
	\caption{AMGA attack for visual tracking. SiamRPN++ accurately locates the target object in the original frame, as shown in (a). Our proposed AMGA attack reduces tracking effectiveness by adding almost imperceptible perturbations, as shown in (b).}
	\label{fig:1}
\end{figure}

To address the abovementioned issues, this paper innovatively proposes the adaptive meta-gradient adversarial attack (AMGA) method, aiming to enhance the effectiveness and transferability of adversarial examples in visual tracking tasks, as shown in Fig.\ref{fig:1}.
AMGA integrates a multi-model ensemble strategy, leveraging the complementarity between different models. Unlike existing multi-model ensemble methods for static image adversarial defense\cite{r40}, we additionally introduce a meta-learning framework\cite{r21}  in the ensemble learning of AMGA, which allows AMGA to dynamically adjust its optimization strategy during the attack process to adapt to the defense mechanisms of visual trackers, further improving attack effectiveness. This approach not only boosts the overall transferability through ensemble learning but also makes the generated adversarial examples more effective in degrading the performance of individual models.
Although the deep learning models used in AMGA, such as InceptionV3\cite{r13}, ResNet-152\cite{r14}, MobileNetV2\cite{r15}, DenseNet-161\cite{r16}, VGG-19\cite{r17}, AlexNet\cite{r18}, SqueezeNet\cite{r19}, and Wide ResNet-50\cite{r20}, are primarily designed for image classification tasks, their feature extraction capabilities have been successfully applied to adversarial attacks in visual tracking tasks through a carefully-designed ensemble strategy. The contributions of this paper primarily include the following aspects:

\begin{itemize}
	\item We propose a black-box adversarial attack method integrating multi-model ensemble strategy and meta-learning framework. The former enables AMGA to simulate various potential attack scenarios. At the same time, the latter allows AMGA to dynamically adjust optimization strategies during the attack process according to the defense mechanisms of the target models, thereby generating adversarial examples that perform effectively across different models.
	\item We utilize momentum mechanisms and Gaussian smoothing to enhance the robustness of adversarial examples. By accumulating historical gradient information, the momentum mechanism reduces fluctuations in gradient updates. Meanwhile, Gaussian smoothing minimizes perturbation noise generated during attacks, enabling the generated adversarial examples to maintain high robustness when handling diverse input images and complex environmental variations.
	\item We conduct extensive validation on seven representative visual trackers, including SiamCAR\cite{r10}, SiamRPN++\cite{r9}, DiMP\cite{r22}, MixFormer\cite{r23}, TransT\cite{r24}, OSTrack\cite{r48}, and SeqTrack\cite{r49}, on three public large-scale datasets: OTB2015\cite{r43}, GOT-10k\cite{r45}, and LaSOT\cite{r44}. Experimental results demonstrate that the proposed AMGA method achieves outstanding attack effectiveness and transferability in black-box scenarios.
\end{itemize}

The rest of this paper is organized as follows: Section \ref{sec:2} reviews recent works relevant to the study. Section \ref{sec:3} details the proposed AMGA attack method. Section \ref{sec:4} presents the experimental analysis and discussions. Finally, Section \ref{sec:5} concludes the paper.

\section{Related works}\label{sec:2}

\subsection{Visual tracking}

Offline and online trackers represent the mainstream visual tracking methods. The formers do not update model parameters during inference, offering higher processing speed. Methods based on Siamese networks have played a significant role in visual tracking. SiamFC\cite{r8}, the pioneer of Siamese tracking methods, reformulates the visual tracking task as a similarity matching problem in a fully convolutional network, achieving efficient tracking without online updates. It uses a shallow AlexNet to extract features from the template and search region, locating the target object by computing their cross-correlation. SiamRPN\cite{r6} builds on this by introducing the region proposal network (RPN) and adding independent classification and regression modules to the Siamese network. SiamRPN predicts candidate regions and their confidence scores while refining the candidate regions for more accurate localization. SiamRPN++\cite{r9} replaces the shallow AlexNet in SiamRPN with a deeper ResNet, significantly improving feature learning capability and tracking performance. It also incorporates refined network designs, such as a feature pyramid and balanced loss functions, allowing it to handle complex and dynamic scenarios more effectively.

Beyond these Siamese network methods, recent research has introduced advanced tracking frameworks.
MixFormer\cite{r23} integrates feature extraction and tracking modules into a hybrid architecture, leveraging the modeling power of Transformers\cite{r27} to greatly improve robustness and accuracy. TransT\cite{r24} introduces a Transformer architecture to implement feature fusion and attention mechanisms, achieving high performance in visual tracking through efficient integration of cross-modal features.
OSTrack\cite{r48} combines feature extraction and relationship modeling, proposing a simple and efficient single-stage tracking method. The specially designed candidate elimination module significantly improves tracking speed and alleviates the interference of redundant background regions on similarity matching. SeqTrack\cite{r49} regards the visual tracking task as a sequence generation problem, predicting bounding boxes in an autoregressive manner. It adopts a simple encoder-decoder Transformer architecture, and this sequence learning paradigm not only simplifies the tracking process but also achieves competitive performance on multiple benchmark datasets.
Compared to traditional Siamese networks, these Transformer-based trackers better capture the complex relationships between the target object and background surroundings, driving further advancements in visual tracking.

Online trackers dynamically adjust model parameters during inference by collecting and analyzing tracking results from previous frames. This process optimizes the representation of the target appearance to address challenges such as illumination variations, occlusion, and deformation, adapting to the dynamic changes in the current frame. UpdateNet\cite{r25} combines online updating strategies with the Siamese network, integrating an update module to maintain the model's timeliness. This method preserves the efficiency of similarity measurement inherent in Siamese networks while enhancing the model's adaptability to target dynamics through online updates, ensuring continuity and accuracy in tracking. By maintaining temporal consistency and coherence in motion trajectories, UpdateNet effectively handles dynamic scenarios. DiMP\cite{r22} introduced an end-to-end optimization framework and a discriminative loss function, leveraging the appearance information of both the target and background for optimal model prediction. PrDiMP\cite{r26} proposed a probabilistic regression framework to effectively address issues caused by label noise, enhancing the stability of the tracker in complex scenarios. This probabilistic approach helps ensure robust performance across diverse and challenging tracking conditions.

\subsection{Adversarial attacks in computer vision}

Adversarial attacks were first introduced in image classification tasks\cite{r2}. The goal is to mislead models into making incorrect decisions by adding carefully crafted and effective perturbations, known as adversarial examples. These perturbations are almost imperceptible to the human vision system. FGSM\cite{r2}, a cornerstone of adversarial attack, operates by calculating the gradient of the loss function to the input image and adding perturbations along the direction of the gradient's sign. This method is computationally simple and efficient, but the adversarial examples it generates are less effective against complex models due to the coarse nature of single-gradient estimation. I-FGSM\cite{r11} was developed to enhance attack success rates. It iteratively updates perturbations, progressively approaching optimal adversarial examples that maximize model loss. This iterative approach generates more precise adversarial examples, thereby improving attack effectiveness. However, these methods primarily target static images and rely on the gradient information of a single model. Consequently, the adversarial examples they produce exhibit poor transferability across models, limiting their applicability in scenarios requiring cross-model attacks.

To solve the problem of adversarial examples performing poorly across multiple models, researchers have proposed techniques such as multi-model ensembles and input diversity. Tramer \et  introduced a method that generates more transferable adversarial examples by integrating multiple substitute models\cite{r28}. This method leverages the diversity of these models in the feature space to create more universal adversarial examples, enhancing the attacking success in black-box scenarios. Additionally, the momentum iterative method (MIM) has been introduced to significantly improve the stability and effectiveness of adversarial examples in cross-model transfer\cite{r30}. By incorporating a momentum term, MIM accumulates the history of gradients, smoothing out fluctuations in the gradient during the iterative process. This reduces instability caused by gradient estimation noise, resulting in more consistent and robust adversarial perturbations. This method improves the efficiency of adversarial example generation and the transferability across different models, ensuring that adversarial attacks maintain high success rates even in unknown or black-box scenarios.

\subsection{Adversarial attacks in visual tracking}

In visual tracking, adversarial attacks primarily focus on generating adversarial examples that mislead the tracker\cite{r31}. Lei \et introduced the ghost adversarial attack, which creates perturbations in the input frames\cite{r32}. This process often incorporates gradient information to modify the feature space of the target object, with the changes being small enough that they do not cause noticeable visual differences. Jia \et focused on generating perturbations for frame-by-frame tracking results by considering motion information\cite{r33}. They first perform a white-box attack by generating time-based perturbations using a known tracker, then transfer them to an unknown tracker in a black-box scenario. This method improves the versatility of the attack by targeting both white-box and black-box trackers. Liang \et proposed a novel approach using drift and embedded feature loss\cite{r34}. Instead of directly attacking the target position prediction, the method adjusts features like similarity in the embedding space, causing the tracker to produce biased target localization predictions.

The above methods demonstrate that, in white-box scenarios, adversarial attacks can effectively degrade tracking performance by precisely calculating gradients. However, in real-world applications, visual tracking models are often part of complex computer vision systems with highly encapsulated internal structures and parameters that are not easily accessible. This severely limits the practical applicability of white-box attacks. As a result, black-box attacks have become a more challenging yet practical direction for research. To address this challenge, several methods have been proposed to enhance the transferability of adversarial examples in black-box scenarios. Zhao \et developed a flexible adversarial attack method, which generates small yet effective perturbations by optimizing an adversarial objective function\cite{r35}. Chen \et introduced the reinforcement adversarial attack, where the attacker is treated as an agent that interacts with the tracker to learn how to generate the most effective perturbations\cite{r36}. Using reinforcement learning, the agent gradually refines its attack strategy, allowing the generated adversarial examples to interfere with a specific tracker and improve the transferability across various trackers. Guo \et explored how adversarial blurring could be used to interfere with visual tracking models\cite{r37}. They proposed an adversarial attack method that applies a blurring effect in the target region, causing the loss of fine details. This, in turn, hampers the tracker’s ability to recognize and localize the target object, making it an effective form of disruption in the tracking process.

\subsection{Adversarial attacks based on meta-learning}

Meta-learning can optimize the generation process of adversarial examples by sharing the learning experience of the model in multiple tasks, making it more adaptable and transferable between different models and tasks. The model-agnostic meta-learning (MAML) method of Finn \et provides a basis for quickly adjusting model parameters in multiple tasks\cite{r21}. It not only provides a new perspective for cross-task generalization of deep learning models but also opens up a new path for the generation of adversarial examples. Through the MAML framework, the generation process of adversarial examples is optimized to be more efficient. These examples can show more robust adaptability and transferability between different models and tasks, effectively coping with complex and changing defense mechanisms. Tramer \et proposed to generate more transferable adversarial examples by attacking multiple alternative models to improve the success rate of attacks in black-box scenarios\cite{r28}. Yuan \et generated adversarial examples that can maintain efficient attacks between different models by integrating models with other structures, further enhancing the effect of black-box attacks\cite{r39}. However, applying meta-learning-based adversarial attacks in dynamic scenarios such as visual tracking is relatively rare, and related research is still in the initial exploratory stage.

In visual tracking, adversarial attacks often face the dual challenges of robustness and transferability. For instance, when the target object is in cases of deformation, illumination variations, background clutter, and occlusion, the generated adversarial examples often fail in these conditions. Particularly in black-box scenarios, one major issue is how to maintain efficient attacks while ensuring the temporal consistency and robustness of adversarial examples. This is a problem that still requires further resolution. Inspired by the multi-model ensemble strategy and meta-learning frameworks used in adversarial example generation, we take a novel approach that deviates from traditional single-model attack methods. By dynamically selecting a subset of models for attack optimization, the transferability of adversarial examples is enhanced. The idea behind this is to utilize the diversity and complementary strengths of different models to improve the robustness and adaptability of adversarial attacks across various scenarios. Furthermore, integrating momentum mechanisms and Gaussian smoothing into the adversarial example generation process further enhances the robustness and deceptive power of these examples. The momentum mechanism helps stabilize adversarial perturbations across iterations by smoothing out noisy gradient updates, while Gaussian smoothing can make perturbations less noticeable, further increasing their ability to deceive the tracker.

\section{Methodology}\label{sec:3}

The proposed AMGA attack method significantly enhances the transferability and robustness of adversarial examples in visual tracking tasks by integrating multiple strategies: meta-learning, multi-model ensemble, momentum mechanism, and Gaussian smoothing.

\subsection{Overview}

\begin{figure*}[t]
	\begin{center}
		\includegraphics[width=\linewidth]{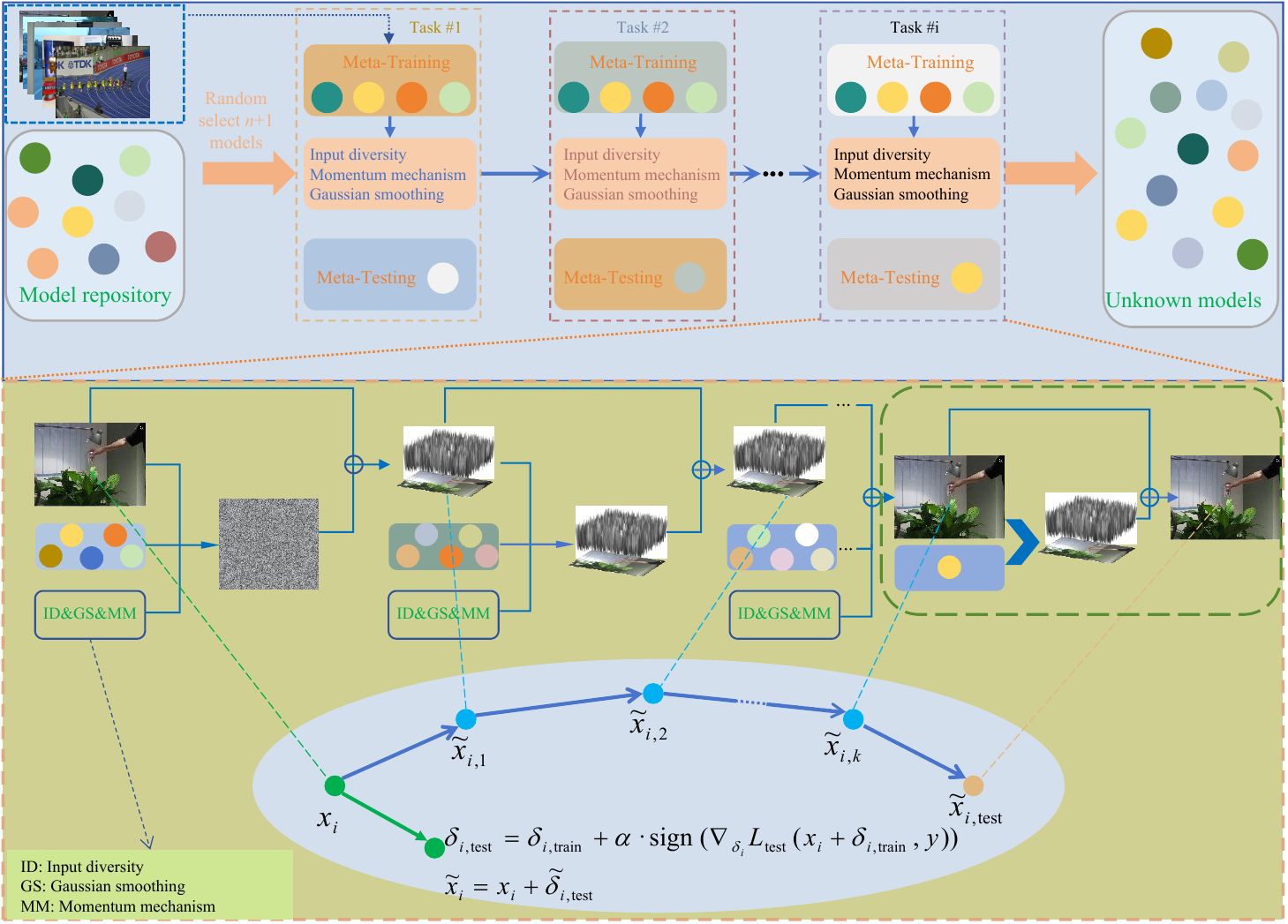}
	\end{center}
	\vspace{-0.8em}
	\caption{Overview pipeline of the proposed AMGA attack method. AMGA randomly selects $n+1$ models from the model repository in each iteration to construct a meta-learning task. Each task is divided into two stages: meta-training and meta-testing. In the meta-training stage, a gradient-based perturbation generation is conducted using the set of the first $n$ models, and this process is iterated $k$ times. In the meta-testing stage, the adversarial perturbation obtained from meta-training is applied to the $(n+1)$-th model to obtain the final adversarial perturbation and example.}
	\label{fig:2}
\end{figure*}

Continuous object detection in visual tracking relies on stable feature extraction and matching. AMGA effectively interferes with the tracker's decision-making process by utilizing high-level features extracted from multiple basic deep learning models, including InceptionV3\cite{r13}, ResNet-152\cite{r14}, MobileNetV2\cite{r15}, DenseNet-161\cite{r16}, VGG-19\cite{r17}, AlexNet\cite{r18}, SqueezeNet\cite{r19}, and Wide ResNet-50\cite{r20}. These basic models focus on different aspects of feature representation due to their varying architectures and training objectives. For example, InceptionV3 excels at capturing multi-scale feature maps, while ResNet152 leverages deep residual networks to extract deeper semantic information. By integrating the features of these models, AMGA can more comprehensively interfere with the feature extraction process, allowing the generated adversarial examples to better represent the target object and improve their adaptability in different scenarios. Specifically, the generated adversarial examples disrupt critical features in the input frame, causing the tracker to fail in correctly extracting features. For example, interference with edge features may prevent the tracker from identifying the object's outline, while interference with texture features may prevent the tracker from matching the object's appearance, leading to tracking failure.

Moreover, AMGA significantly enhances the transferability of adversarial examples through the multi-model ensemble\cite{r40}. In practical applications, attackers often cannot access the specific structure and parameters of the target model, thus operating in black-box scenarios. Traditional adversarial attack methods rely on a single model and often generate adversarial examples that perform poorly on unknown models. However, adversarial examples generated by AMGA, optimized across multiple models, exhibit more robust generalization capabilities. This ensures that even in black-box scenarios, where the attacker is unaware of the model's specific details, these adversarial examples remain effective across various models, addressing the limitations of traditional attack methods in terms of transferability. Additionally, integrating the outputs of multiple models helps balance the performance fluctuations that may arise in a single model, allowing the generated adversarial examples to maintain more stable attack effectiveness when faced with real-world factors such as deformation, occlusion, and motion blur. The overview pipeline of the proposed AMGA attack method is shown in Fig.\ref{fig:2}.

\subsection{Meta-training}

AMGA divides the generation process of adversarial examples into meta-training and meta-testing, and combines the multi-model ensemble strategy to enhance the generalization and transferability of adversarial examples.

During the meta-training stage, a subset $M_{\text{train}}=\{m_1,m_2,\ldots,m_n\}$ containing $n$ deep learning models is randomly selected from a repository $M=\{m_1,m_2,\ldots,m_N\}$ of $N$ pre-trained models to generate adversarial examples, \ie $M_{\text{train}}\subseteq M$. The repository $M$ can include widely used CNN architectures. The composition of the subset $M_{\text{train}}$ is random and simulates various possible attack scenarios, thereby improving the generalization capability of adversarial examples and ensuring their robustness across different model architectures.

For a given input frame $x$, the output of each model $m_i$ is a class probability distribution vector $p_i(x)$, representing the model's confidence in each class. We use a weighted summation scheme to merge the outputs of multiple models, resulting in a new ensemble probability as,
\begin{equation}\label{eq:1}
	p_{\text{ensemble}}(x)=\sum_{i=1}^n\beta_i\cdot p_i(x),
\end{equation}
where $\beta_i$ represents the weight coefficient of the $i$-th model, satisfying $\sum_{i=1}^n\beta_i=1$. We allow $\beta_i$ to be learned during gradient descent, enabling the automatic determination of the optimal weight allocation to enhance attack effectiveness and transferability.

Next, we compute the ensemble loss based on the combined probability and use the gradient of this loss to generate adversarial perturbations. The ensemble loss is defined as,
\begin{equation}\label{eq:2}
	\mathcal{L}_{\text{ensemble}}(x,y)=\mathbb{E}_{(x,y)\sim\mathbb{D}}[\mathcal{L}(p_{\text{ensemble}}(x),y)],
\end{equation}
where $\mathcal{L}$ represents the cross-entropy loss, which measures the difference between the ensemble probability distribution $p_{\text{ensemble}}(x)$ and the true label $y$ for the input frame $x$ from the dataset $\mathbb{D}$.

Finally, the gradient of the ensemble loss is calculated to determine the updated direction of the adversarial perturbation $\delta$. In the $(k+1)$-th iteration, the adversarial perturbation $\delta$ is updated using the gradient of the ensemble loss as,
\begin{equation}\label{eq:3}
	\delta_{k+1}=\delta_k+\alpha\cdot\mathrm{sign}(\nabla_\delta\mathcal{L}_{\text{ensemble}}(x+\delta_k,y)),
\end{equation}
where $\alpha$ represents the update step size of the adversarial perturbation $\delta$, and $\mathrm{sign}(\nabla_\delta\mathcal{L}_{\text{ensemble}}(x+\delta_k,y))$ ensures $\delta$ is updated towards the direction that maximizes the ensemble loss, so that the generated adversarial examples can achieve effective attacks for multi-models.

\subsection{Momentum mechanism}

To reduce oscillations during the gradient update process, we introduce the momentum mechanism\cite{r30}, which dynamically adjusts the gradient update direction based on the historical information of the accumulated gradients in each iteration, reducing the instability caused by randomly selecting models.
By dynamically adjusting the perturbation generation process, we ensure that the generated adversarial examples maintain effective attack capability across different models and input frames. The momentum mechanism can be expressed as,
\begin{equation}\label{eq:4}
	m_{k+1}=\mu\cdot m_k+\frac{\nabla_\delta\mathcal{L}_{\text{ensemble}}(x+\delta_k,y)}{\|\nabla_\delta\mathcal{L}_{\text{ensemble}}(x+\delta_k,y)\|},
\end{equation}
where $m_k$ represents the accumulated gradient at the $k$-th iteration, and $\mu$ is the momentum coefficient, which balances the contributions of the current gradient and the gradients from the previous iterations.

By incorporating the momentum mechanism, the accumulated gradient helps stabilize the update direction, reducing instability caused by large gradient variations. Therefore, Eq.\ref{eq:3} can be rewritten as,
\begin{equation}\label{eq:5}
	\delta_{k+1}=\delta_k+\alpha\cdot\mathrm{sign}(m_{k+1}),
\end{equation}

\subsection{Gaussian smoothing}

We further applied Gaussian smoothing to reduce the high-frequency components of adversarial examples, enhancing their transferability and attack effectiveness across different models. The Gaussian smoothing can be expressed as,
\begin{equation}\label{eq:6}
	\tilde{\delta}=\delta * \mathcal{G}_\sigma,
\end{equation}
where $\tilde{\delta}$ is the smoothed perturbation after applying Gaussian smoothing, $\delta$ represents the original adversarial perturbation, $*$ denotes the convolution operation, and $\mathcal{G}_\sigma$ is the Gaussian smoothing operation, which is defined as,
\begin{equation}\label{eq:7}
	\mathcal{G}_\sigma(u,v)=\frac{1}{2\pi\sigma^2}\exp(-\frac{u^2+v^2}{2\sigma^2}),
\end{equation}
where $(u,v)$ are the coordinates of the kernel in the 2D space, $\sigma$ is the standard deviation of the Gaussian kernel, controlling the extent of smoothing. The smoothed perturbation improves transferability across different models, ensuring that the adversarial example retains its efficacy while appearing more natural and less detectable. This combination of smoothing and ensemble integration optimizes both the stealthiness and robustness of the attack.

\subsection{Meta-testing}

During the meta-testing stage, the generated adversarial examples are used to evaluate their attack effectiveness on unseen models. To verify the transferability of adversarial samples, an unseen model $m_{\text{test}}$ is selected from the model repository $M$, which was not used in the meta-training stage. The loss of the adversarial examples is calculated using the output of the unseen model, and the perturbations are further optimized. The meta-testing loss is defined as,
\begin{equation}\label{eq:8}
	\mathcal{L}_{\text{test}}(x+\delta_{\text{train}},y)=\mathbb{E}_{(x,y)\sim\mathbb{D}}[\mathcal{L}(p_{\text{test}}(x+\delta_{\text{train}}),y)],
\end{equation}
where $\delta_{\text{train}}$ is the final training perturbation obtained after all iterations are completed in the meta-training stage. We further update the perturbation by calculating the gradient of the meta-testing loss,
\begin{equation}\label{eq:9}
	\delta_{\text{test}}=\delta_{\text{train}}+\alpha\cdot\mathrm{sign}(\nabla_\delta\mathcal{L}_{\text{test}}(x+\delta_{\text{train}},y)),
\end{equation}
where $\nabla_\delta\mathcal{L}_{\text{test}}$ is the gradient of the meta-testing loss for the perturbation.

After that, the adversarial example $\tilde{x}$ can be obtained as,
\begin{equation}\label{eq:10}
	\tilde{x}=x+\tilde{\delta}_{\text{test}}
\end{equation}

The optimization process in the meta-testing stage ensures the effectiveness of adversarial examples on unseen models, significantly improving their transferability.
By continuously alternating between the meta-training and meta-testing processes, the adversarial examples are encouraged to target the common features across different models. Even when there is a significant difference between black-box and white-box scenarios, AMGA can still enhance the robustness of the adversarial examples across different models through various strategies, ensuring stable performance in both unpredictable black-box and white-box scenarios. Specifically, the multi-model ensemble learning used by AMGA covers a broader feature space, while the momentum mechanism suppresses gradient fluctuations, enhancing the stability of the perturbations. Therefore, the final adversarial examples effectively attack the models used in meta-training and maintain a high transferability against previously unseen models in meta-testing.

\section{Experiments}\label{sec:4}

To validate the attack performance of AMGA in visual tracking, we conducted experiments on multiple mainstream trackers and widely used benchmark datasets. Specifically, we selected seven representative trackers with different architectures: SiamCAR\cite{r10}, SiamRPN++\cite{r9}, DiMP\cite{r22}, MixFormer\cite{r23}, TransT\cite{r24}, OSTrack\cite{r48}, and SeqTrack\cite{r49}, and tested the generated adversarial examples on three benchmark datasets: OTB2015\cite{r43}, GOT-10k\cite{r45}, and LaSOT\cite{r44}.

\subsection{Implementation details}

We selected seven different types of visual trackers to validate the generality of our black-box adversarial attacks: (a) SiamCAR\footnote{\url{https://github.com/ohhhyeahhh/SiamCAR}.} directly predicts the object class probabilities and bounding box positions through a fully convolutional network, avoiding the complex region proposal scheme; (b) SiamRPN++\footnote{\url{https://github.com/STVIR/pysot}.} is an improved version of SiamRPN that replaces the shallow AlexNet backbone with a deeper architecture ResNet; (c) DiMP\footnote{\url{https://github.com/visionml/pytracking}.} is a dynamic model prediction tracker capable of adaptively updating itself based on changes in target and background appearances; (d) MixFormer\footnote{\url{https://github.com/MCG-NJU/MixFormer}.} improves tracking performance by integrating multiple features, making it well-suited for diverse target objects and background surroundings; (e) TransT\footnote{\url{https://github.com/chenxin-dlut/TransT}.} captures long-range dependencies of the target for precise localization based on the Transformer architecture; (f) OSTrack\footnote{\url{https://github.com/botaoye/OSTrack}.} combines feature extraction and relationship learning, performing exceptionally well in complex scenarios; and (g) SeqTrack\footnote{\url{https://github.com/microsoft/VideoX/tree/master/SeqTrack}.} designs a multimodal sequence-to-sequence learning framework, achieving robust object localization. With their varied designs and strengths, these trackers provide a comprehensive platform to evaluate the effectiveness and generality of the proposed adversarial attack across varying architectures and scenarios. It should be noticed that all models we used were straightforwardly downloaded from public code repositories, and all model parameters remained consistent with those released by the authors without any special fine-tuning.

All the experiments were conducted on a server with an Intel$^\circledR$ Xeon$^\circledR$ E5-2680 v4 CPU @ 2.4GHz CPU with 128GB RAM, and a NVIDIA$^\circledR$ GeForce RTX$^{\mathrm{TM}}$ 3090 Ti GPU with 24GB VRAM. During the meta-training stage, multiple models were randomly selected from the model repository. Adversarial perturbations were generated based on the loss gradients from these models. The update step size $\alpha$ was set to 0.01 and dynamically adjusted based on optimization progress. The momentum coefficient $\mu$ was set to 0.9 to accelerate convergence and suppress gradient oscillations. In the meta-testing stage, models that were not involved in the training were chosen to validate the attack effectiveness of the adversarial examples. Further optimization of the perturbations was carried out to ensure that the adversarial examples maintained high attack effectiveness on unseen models, thus enhancing their transferability and robustness. The generated adversarial examples were then used on the initial frame, disrupting the tracker's feature extraction and decision-making process.

Additionally, to improve the robustness of the generated adversarial examples, we introduced input diversity in AMGA. This ensured the adversarial examples could maintain a stable attack performance across different input forms and optimization processes. In each iteration, random scaling and padding operations were applied to the input frames to prevent the adversarial examples from relying on a specific tracking scenario and improve the robustness of the adversarial examples across different tracking scenarios.

\subsection{Overall Results}

\begin{table*}[t!]
	\centering
	\caption{Comparison of tracking results with original sequences, random noise, and AMGA attack of SiamCAR\cite{r10}, SiamRPN++\cite{r9}, DiMP\cite{r22}, MixFormer\cite{r23}, TransT\cite{r24}, OSTrack\cite{r48}, and SeqTrack\cite{r49} on the OTB2015 benchmark dataset\cite{r43}. The best values are highlighted in \textbf{bold}.}
	\label{tab:1}
		\begin{tabular}{lccclccc}
			\toprule
			\multirow{2}{*}{Trackers} & \multicolumn{3}{c}{Success rate $\downarrow$}         &  & \multicolumn{3}{c}{Precision $\downarrow$}         \\ \cline{2-4} \cline{6-8}
			& Orig. & Random   & AMGA (Ours)           &  & Orig. & Random   & AMGA (Ours)          \\ \hline
			SiamCAR\cite{r10}                   & 0.673 & 0.645 & \textbf{0.533} &  & 0.880 & 0.847 & \textbf{0.719} \\
			SiamRPN++\cite{r9}                  & 0.639 & 0.626 & \textbf{0.420} &  & 0.849 & 0.827 & \textbf{0.562} \\
			DiMP\cite{r22}                      & 0.664 & 0.648 & \textbf{0.363} &  & 0.860 & 0.844 & \textbf{0.461} \\
			MixFormer\cite{r23}                 & 0.690 & 0.639 & \textbf{0.388} &  & 0.913 & 0.849 & \textbf{0.507} \\
			TransT\cite{r24}                    & 0.678 & 0.648 & \textbf{0.464} &  & 0.874 & 0.839 & \textbf{0.608} \\
			OSTrack\cite{r48}                    & 0.676 & 0.592 & \textbf{0.374} &  & 0.877 & 0.778 & \textbf{0.508} \\
			SeqTrack\cite{r49}                    & 0.687 & 0.623 & \textbf{0.386} &  & 0.895 & 0.822 & \textbf{0.522} \\
			\bottomrule
		\end{tabular}%
\end{table*}

\begin{figure*}[t!]
	\begin{center}
		\includegraphics[width=\linewidth]{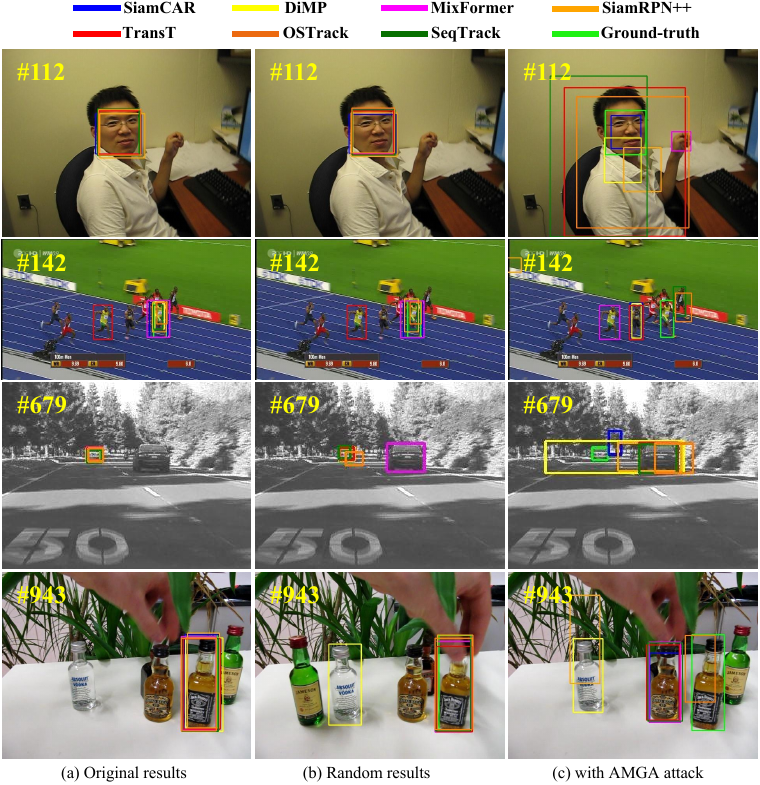}
	\end{center}
	\vspace{-0.8em}
	\caption{Visualization of predicted bounding boxes of SiamCAR\cite{r10}, SiamRPN++\cite{r9}, DiMP\cite{r22}, MixFormer\cite{r23}, TransT\cite{r24}, OSTrack\cite{r48}, and SeqTrack\cite{r49} on original frames, random noised examples, and AMGA adversarial examples from the OTB2015 benchmark dataset\cite{r43}. Ground-truth bounding boxes are also provided for comparison.}
	\label{fig:3}
\end{figure*}

\textbf{Results on OTB2015:} OTB2015\cite{r43} is a classic benchmark dataset for visual tracking, containing 100 fully annotated video sequences used to evaluate tracking performance under different conditions. We selected precision and success rate as evaluation metrics: the former measures the Euclidean distance between the centroid of the predicted bounding box and the ground truth bounding box (typically within 20 pixels), while the latter calculates the proportion of frames where the intersection over union (IoU) between the predicted bounding box and the ground truth bounding box exceeds a predefined threshold. The adversarial examples generated by AMGA significantly impacted both precision and success rates across all trackers. For comparison, we also conducted experiments injecting random noise following Gaussian distribution. The results in Table \ref{tab:1} show that AMGA perturbations are far more effective at degrading performance than random noise. Fig.\ref{fig:3} visualizes the tracking results of the seven selected trackers. With our AMGA attack, the precision of SiamCAR dropped from 88.0\% to 71.9\%, and the success rate decreased from 67.3\% to 53.3\%. For TransT, precision fell from 87.4\% to 60.8\%, and success rate dropped from 67.8\% to 46.4\%. Similar trends were observed for other trackers like DiMP, MixFormer, OSTrack, SeqTrack, and SiamRPN++, demonstrating that AMGA effectively disrupts tracking performance.

\begin{table*}[t!]
	\centering
	\caption{Comparison of tracking results with original sequences, random noise, and AMGA attack of SiamCAR\cite{r10}, SiamRPN++\cite{r9}, DiMP\cite{r22}, MixFormer\cite{r23}, TransT\cite{r24}, OSTrack\cite{r48}, and SeqTrack\cite{r49} on the LaSOT benchmark dataset\cite{r44}. The best values are highlighted in \textbf{bold}.}
	\label{tab:2}
	\resizebox{\textwidth}{!}{%
		\begin{tabular}{lccclccclccc}
			\toprule
			\multirow{2}{*}{Trackers} & \multicolumn{3}{c}{Success rate $\downarrow$} &  & \multicolumn{3}{c}{Precision $\downarrow$} &  & \multicolumn{3}{c}{Norm. Precision $\downarrow$} \\ \cline{2-4} \cline{6-8} \cline{10-12}
			& Orig.    & Random   & AMGA (Ours)   &  & Orig.    & Random   & AMGA (Ours)   &  & Orig.     & Random   & AMGA (Ours)    \\ \hline
			SiamCAR\cite{r10}                & 0.457    & 0.428   & \textbf{0.224}   &  & 0.538    & 0.472    & \textbf{0.262}    &  & 0.485        & 0.391    & \textbf{0.249}      \\
			SiamRPN++\cite{r9}               & 0.495    & 0.494   & \textbf{0.283}   &  & 0.570    & 0.568    & \textbf{0.481}    &  & 0.488        & 0.481    & \textbf{0.227}      \\
			DiMP\cite{r22}                   & 0.569    & 0.513   & \textbf{0.371}   &  & 0.637    & 0.552    & \textbf{0.388}    &  & 0.536        & 0.453    & \textbf{0.314}      \\
			MixFormer\cite{r23}              & 0.702    & 0.682   & \textbf{0.443}   &  & 0.801    & 0.774    & \textbf{0.468}    &  & 0.764        & 0.738    & \textbf{0.418}      \\
			TansT\cite{r24}                  & 0.642    & 0.641   & \textbf{0.458}   &  & 0.730    & 0.718    & \textbf{0.489}    &  & 0.682        & 0.670    & \textbf{0.426}      \\
			OSTrack\cite{r48}                  & 0.691    & 0.678   & \textbf{0.492}   &  & 0.786    & 0.763    & \textbf{0.541}    &  & 0.752        & 0.730    & \textbf{0.486}      \\
			SeqTrack\cite{r49}                  & 0.698    & 0.658   & \textbf{0.430}   &  & 0.797    & 0.753    & \textbf{0.461}    &  & 0.762        & 0.707    & \textbf{0.419}      \\
			\bottomrule
		\end{tabular}%
	}
\end{table*}

\begin{figure*}[t!]
	\begin{center}
		\includegraphics[width=\linewidth]{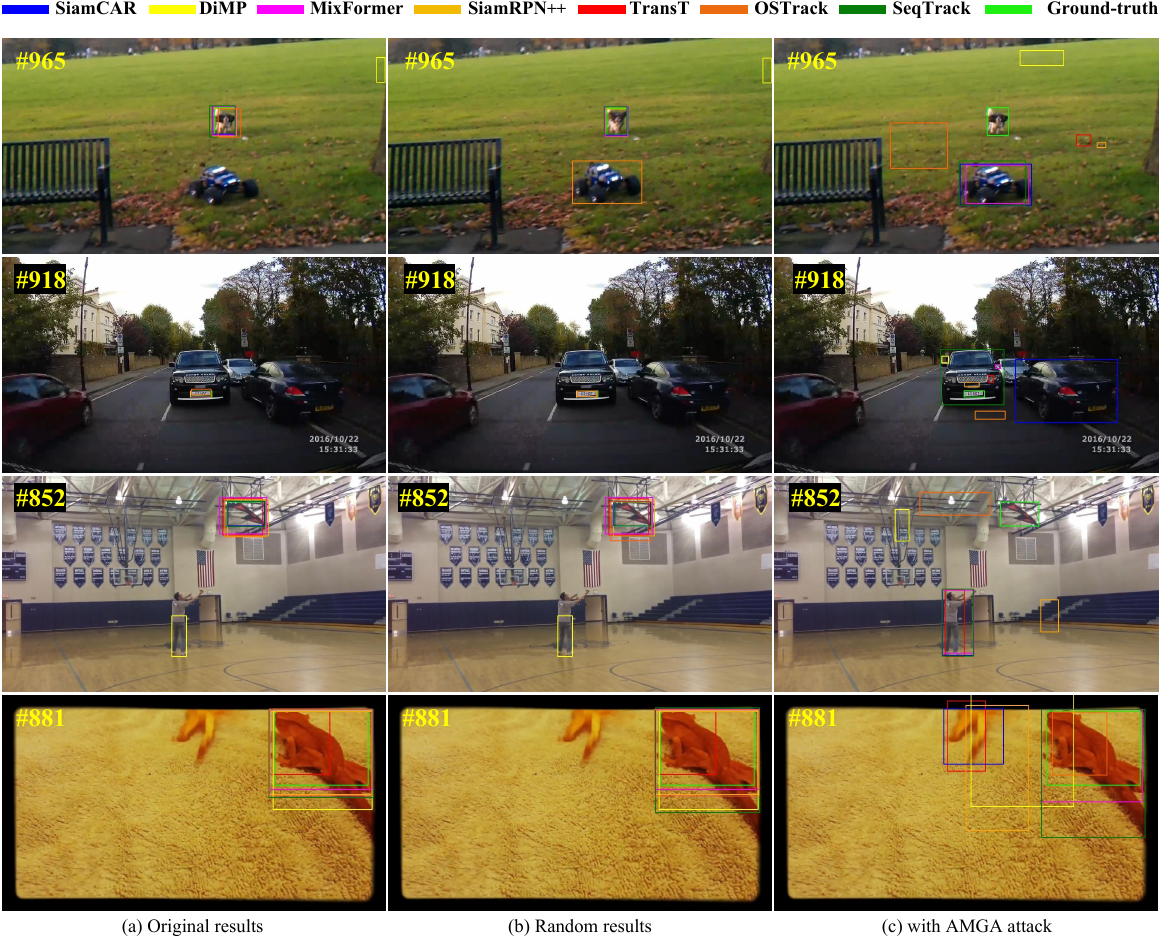}
	\end{center}
	\vspace{-0.8em}
	\caption{Visualization of predicted bounding boxes of SiamCAR\cite{r10}, SiamRPN++\cite{r9}, DiMP\cite{r22}, MixFormer\cite{r23}, TransT\cite{r24}, OSTrack\cite{r48}, and SeqTrack\cite{r49} on original frames, random noised examples, and AMGA adversarial examples from the LaSOT benchmark dataset\cite{r44}. Ground-truth bounding boxes are also provided for comparison.}
	\label{fig:6}
\end{figure*}

\textbf{Results on LaSOT:} LaSOT is a large-scale visual tracking benchmark dataset containing multiple long-term annotated video sequences. It is commonly used to evaluate a tracker’s robustness and generalization capability in practical applications. We selected norm precision, precision, and success rate as the primary evaluation metrics. Precision and success rate are consistent with their definitions in OTB. Norm precision is a normalized version of precision that accounts for target displacement and motion range. Mitigating biases introduced by rapid motion or scale changes provide a more comprehensive assessment of the tracker’s accuracy and robustness. Table \ref{tab:2} presents the performance of the seven selected trackers on the LaSOT dataset, and Fig.\ref{fig:6} visualizes the tracking results. After applying AMGA attacks to these trackers, all metrics showed significant declines, indicating that the adversarial examples generated by AMGA effectively disrupt the accuracy and robustness of these trackers.

\begin{table*}[t!]
	\centering
	\caption{Comparison of tracking results with original sequences, random noise, and AMGA attack of SiamCAR\cite{r10}, SiamRPN++\cite{r9}, DiMP\cite{r22}, MixFormer\cite{r23}, TransT\cite{r24}, OSTrack\cite{r48}, and SeqTrack\cite{r49} on the GOT-10k benchmark dataset\cite{r45}. The best values are highlighted in \textbf{bold}.}
	\label{tab:3}
	\resizebox{\textwidth}{!}{%
		\begin{tabular}{lccclccclccc}
			\toprule
			\multirow{2}{*}{Trackers} & \multicolumn{3}{c}{AO $\downarrow$} &  & \multicolumn{3}{c}{SR@0.5 $\downarrow$} &  & \multicolumn{3}{c}{SR@0.75 $\downarrow$} \\ \cline{2-4} \cline{6-8} \cline{10-12}
			& Orig.    & Random   & AMGA (Ours)   &  & Orig.    & Random   & AMGA (Ours)   &  & Orig.     & Random   & AMGA (Ours)    \\ \hline
			SiamCAR\cite{r10}                & 0.568    & 0.523   & \textbf{0.419}   &  & 0.666    & 0.593    & \textbf{0.479}    &  & 0.425        & 0.374    & \textbf{0.300}      \\
			SiamRPN++\cite{r9}               & 0.512    & 0.481   & \textbf{0.403}   &  & 0.610    & 0.543    & \textbf{0.461}    &  & 0.324        & 0.274    & \textbf{0.187}      \\
			DiMP\cite{r22}                   & 0.606    & 0.512   & \textbf{0.424}   &  & 0.729    & 0.617    & \textbf{0.479}    &  & 0.461        & 0.307    & \textbf{0.266}      \\
			MixFormer\cite{r23}              & 0.720    & 0.682   & \textbf{0.548}   &  & 0.809    & 0.706    & \textbf{0.591}    &  & 0.705        & 0.584    & \textbf{0.466}      \\
			TansT\cite{r24}                  & 0.723    & 0.623   & \textbf{0.586}   &  & 0.824    & 0.691    & \textbf{0.646}    &  & 0.682        & 0.591    & \textbf{0.505}      \\
			OSTrack\cite{r48}              & 0.708    & 0.645   & \textbf{0.491}   &  & 0.801    & 0.701    & \textbf{0.512}    &  & 0.699        & 0.595    & \textbf{0.465}      \\
			SeqTrack\cite{r49}                  & 0.745    & 0.673   & \textbf{0.512}   &  & 0.834    & 0.684    & \textbf{0.602}    &  & 0.713        & 0.584    & \textbf{0.472}      \\
			\bottomrule
		\end{tabular}%
	}
\end{table*}

\textbf{Results on GOT-10k:} GOT-10k is a large-scale visual tracking benchmark dataset that includes diverse object categories and tracking scenarios, designed to evaluate the generalization ability of trackers. We used three evaluation metrics: average overlap (AO) measures the average IoU between predicted and ground truth bounding boxes across the entire test sequence, SR@0.5 and SR@0.75 represent the proportions of frames where the success rate exceeds 0.5 and 0.75, respectively. Table \ref{tab:3} shows that all trackers performed well on the original sequences, but their performance significantly deteriorated under AMGA attacks. This demonstrates the effectiveness of AMGA in disrupting the tracking performance of different trackers.

Experiments on the OTB, LaSOT, and GOT-10k benchmark datasets demonstrate that the adversarial examples generated by AMGA exhibit significant attack effectiveness across different trackers, effectively reducing their precision and success rates. This indicates that AMGA possesses strong transferability and robustness, achieving consistent attack performance across diverse scenarios and object categories.

\subsection{Ablation studies}

\begin{table*}[t!]
	\centering
	\caption{Ablation studies of each component on OTB2015 benchmark dataset\cite{r43} illustrate the attack effectiveness of AMGA on SiamRPN++\cite{r9} under different configurations. The best values are highlighted in \textbf{bold}.}
	\label{tab:4}
		\begin{tabular}{lcccc}
			\toprule
			SiamRPN++\cite{r9}        & Success rate $\downarrow$ & Success drop $\uparrow$ & Precision $\downarrow$ & Precision drop $\uparrow$      \\ \hline
			w/o attack                      & 0.639    & 0.000    & 0.849 & 0.000 \\
			w/ random noise                   & 0.626    & 0.013    & 0.827 & 0.022 \\
			w/ input diversity                & 0.545    & 0.094    & 0.717 & 0.132 \\
			w/ meta-gradient learning         & 0.507    & 0.132    & 0.626 & 0.223 \\
			w/ full AMGA                           & \textbf{0.420}    & \textbf{0.219}    & \textbf{0.562} & \textbf{0.287} \\
			w/o momentum mechanism  & 0.494    & 0.144    & 0.641& 0.208 \\
			w/o Gaussian smoothing          & 0.485    & 0.154    & 0.658 & 0.191 \\
			\bottomrule
		\end{tabular}%
\end{table*}

\begin{table*}[t!]
	\centering
	\caption{Ablation study of the effectiveness of the AMGA attack on SiamRPN++\cite{r9} on the OTB2015 benchmark dataset\cite{r43} using Gaussian smoothing with different standard deviations. The best values are highlighted in \textbf{bold}.}
	\label{tab:6}
		\begin{tabular}{ccccc}
			\toprule
			Standard deviation ($\sigma$)        &  Success drop $\uparrow$ & Precision drop $\uparrow$  & PSNR (dB) $\uparrow$ & SSIM $\uparrow$   \\ \hline
			0.5                & \textbf{0.259}    & \textbf{0.353}    & 26.29 & 0.851 \\
			1.0                & 0.219    & 0.287    & \textbf{31.76} & \textbf{0.906} \\
			2.0                & 0.166    & 0.219    & 27.81 & 0.883 \\
			\bottomrule
		\end{tabular}%
\end{table*}

\begin{table*}[t!]
	\centering
	\caption{Comparison of effectiveness and efficiency with existing white-box and black-box adversarial attack methods for SiamRPN++\cite{r9} on the OTB2015 benchmark dataset\cite{r43}. The best values are highlighted in \textbf{bold}.}
	\label{tab:5}
		\begin{tabular}{lcccc}
			\toprule
			Attack methods        & Success drop $\uparrow$ & Precision drop $\uparrow$ & Speed (fps) $\uparrow$ & Type      \\ \hline
			CSA\cite{r46}           & 0.372    & 0.443   & 2.6 & White-box \\
			SPARK\cite{r47}         & 0.066     & 0.027   & \textbf{3.5}  & Black-box \\
			IoU\cite{r12}           & 0.196    & 0.261    & 1.5 & Black-box \\
			AMGA (Ours)             & \textbf{0.219}    & \textbf{0.287}   & 1.1 & Black-box \\
			\bottomrule
		\end{tabular}%
\end{table*}

\begin{figure*}[t!]
	\begin{center}
		\subfigure{\includegraphics[width=.328\linewidth]{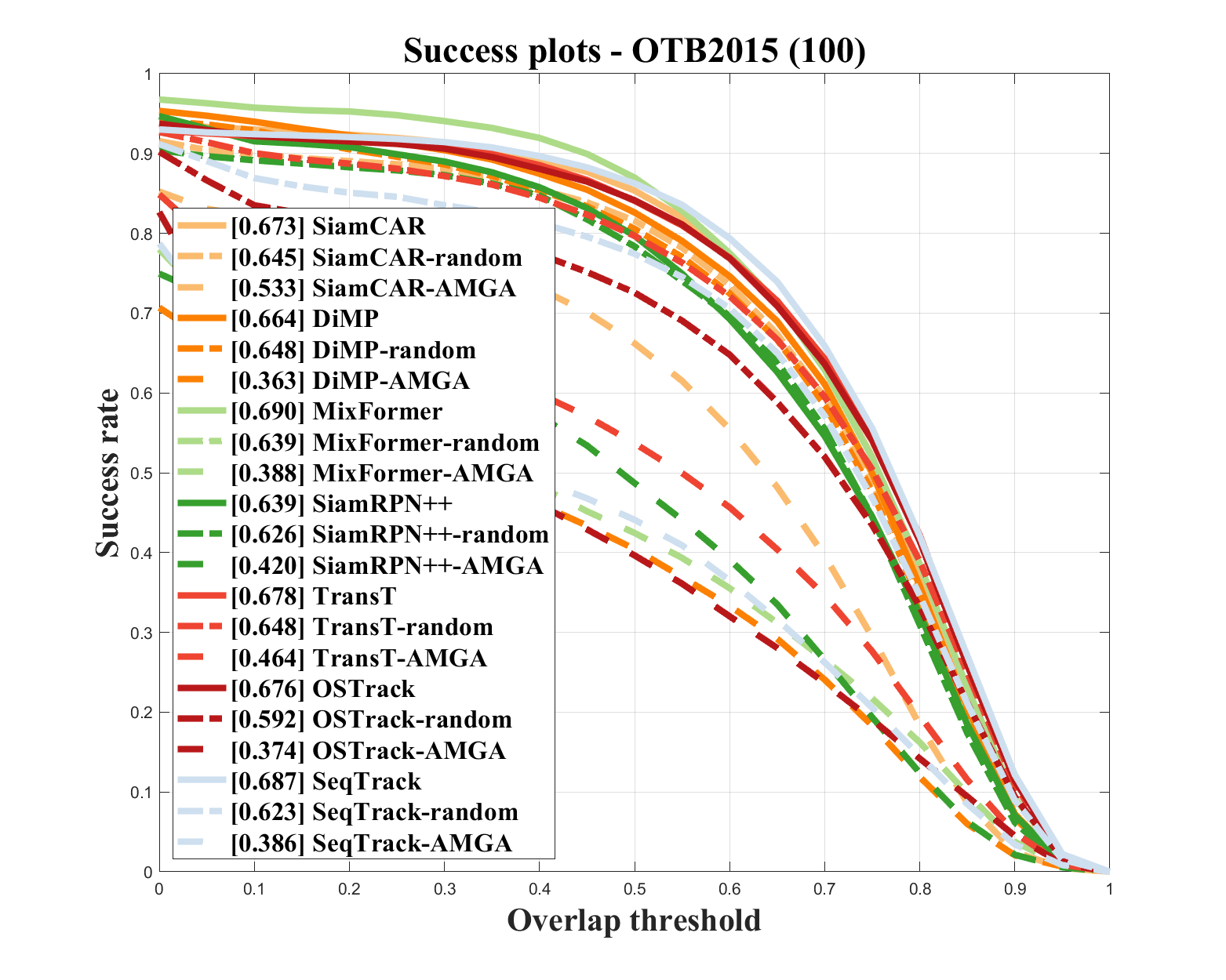}}
		\hspace{0.05em}
		\subfigure{\includegraphics[width=.328\linewidth]{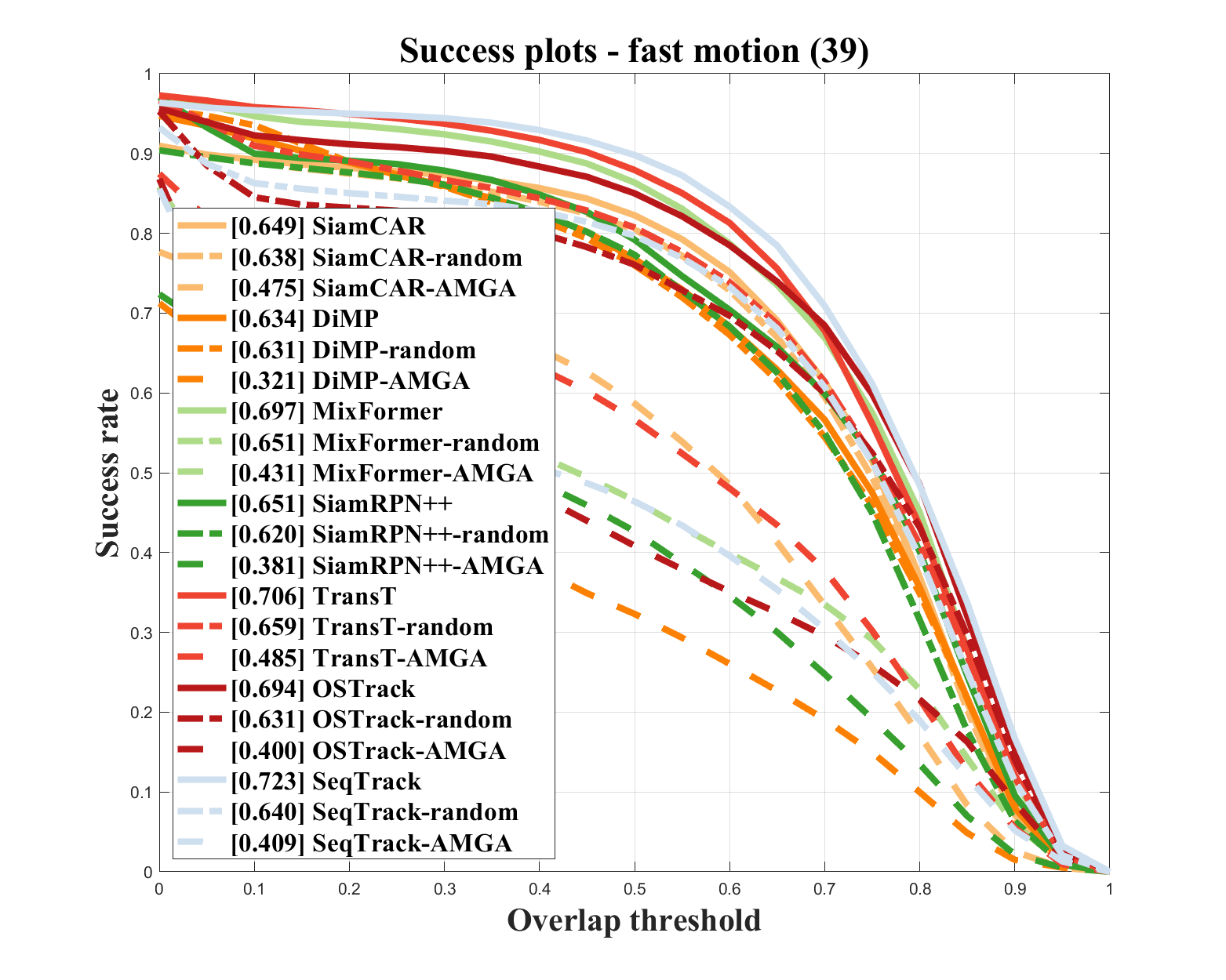}}
		\hspace{0.05em}
		\subfigure{\includegraphics[width=.328\linewidth]{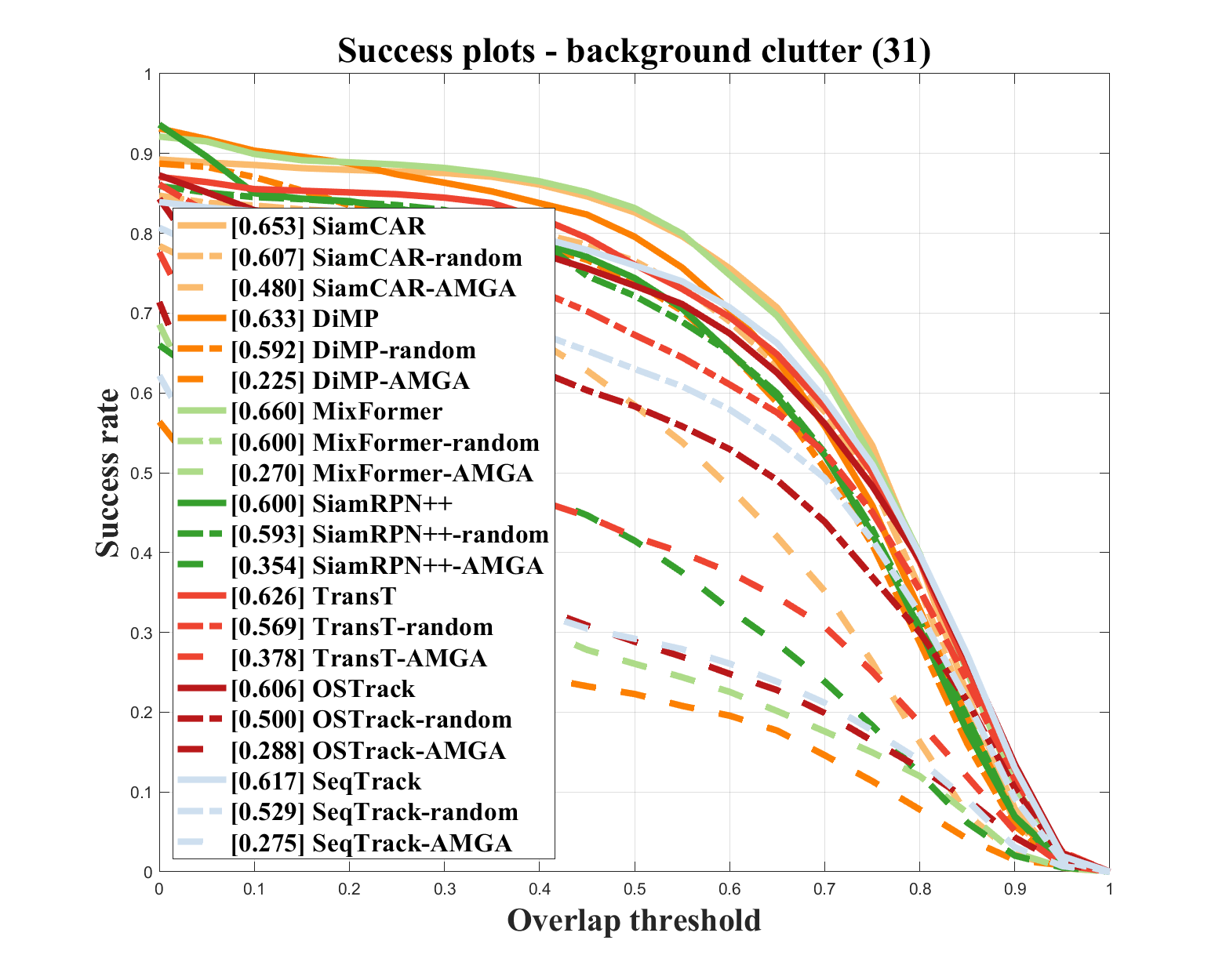}}
		\vfill
		\subfigure{\includegraphics[width=.328\linewidth]{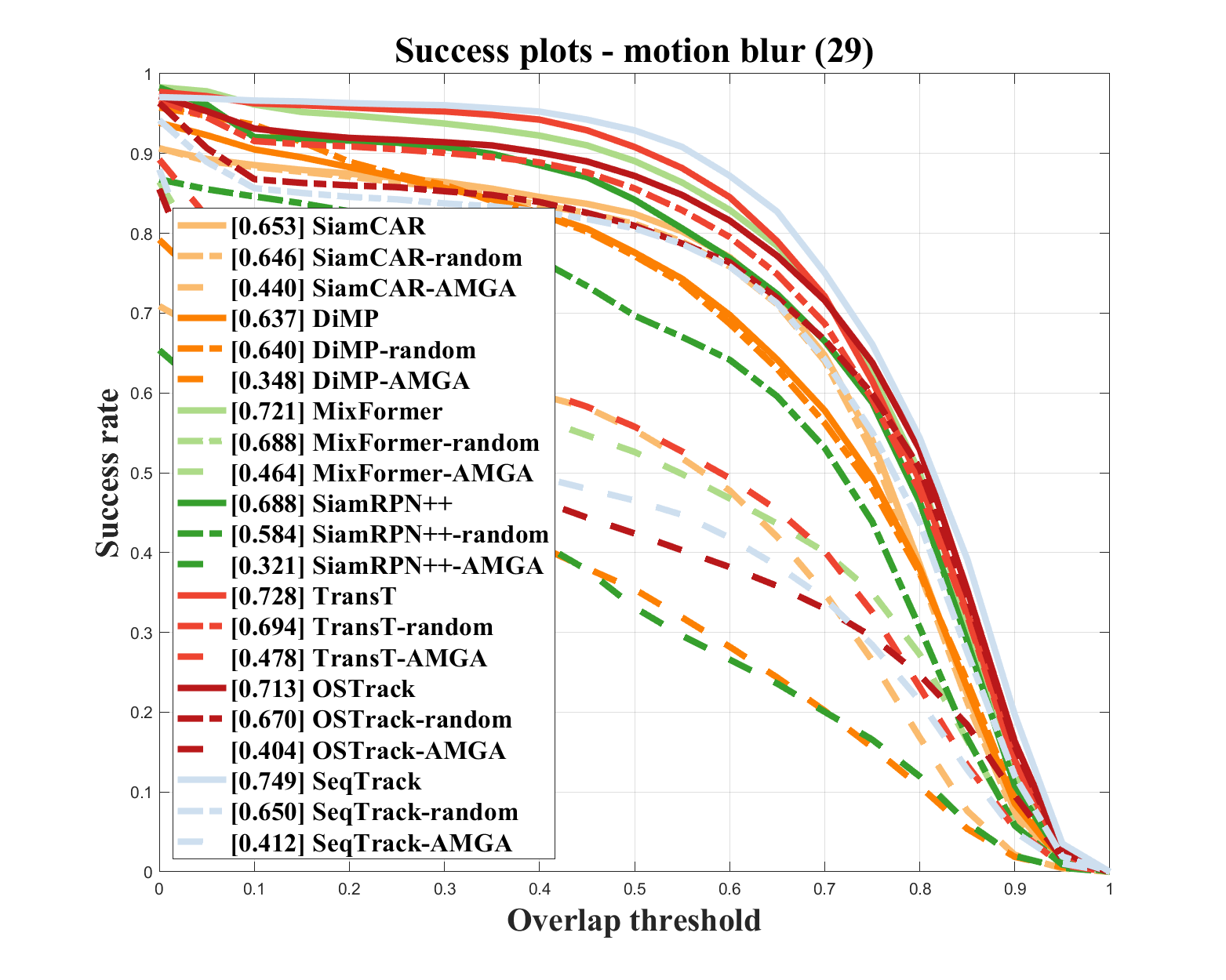}}
		\hspace{0.05em}
		\subfigure{\includegraphics[width=.328\linewidth]{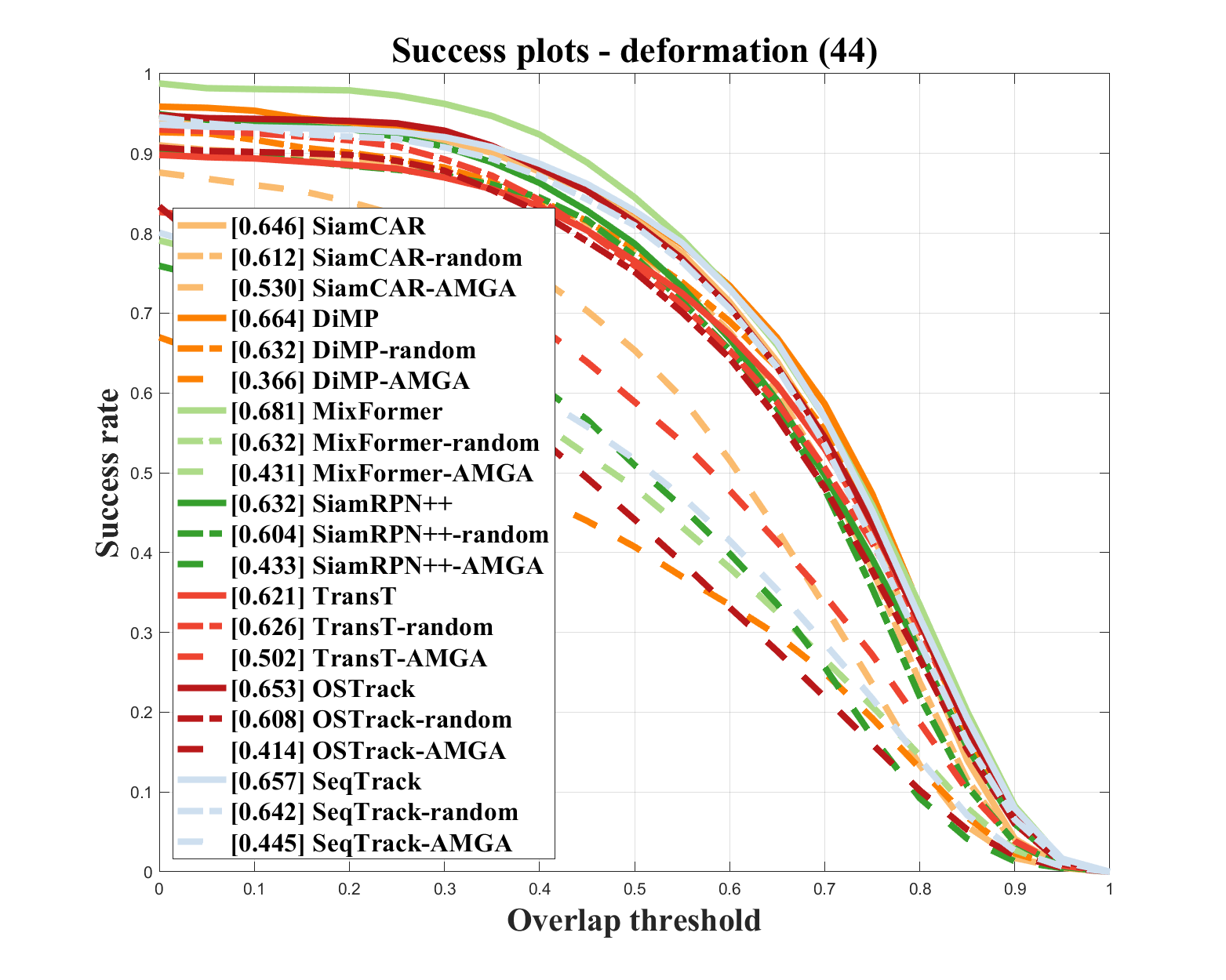}}
		\hspace{0.05em}
		\subfigure{\includegraphics[width=.328\linewidth]{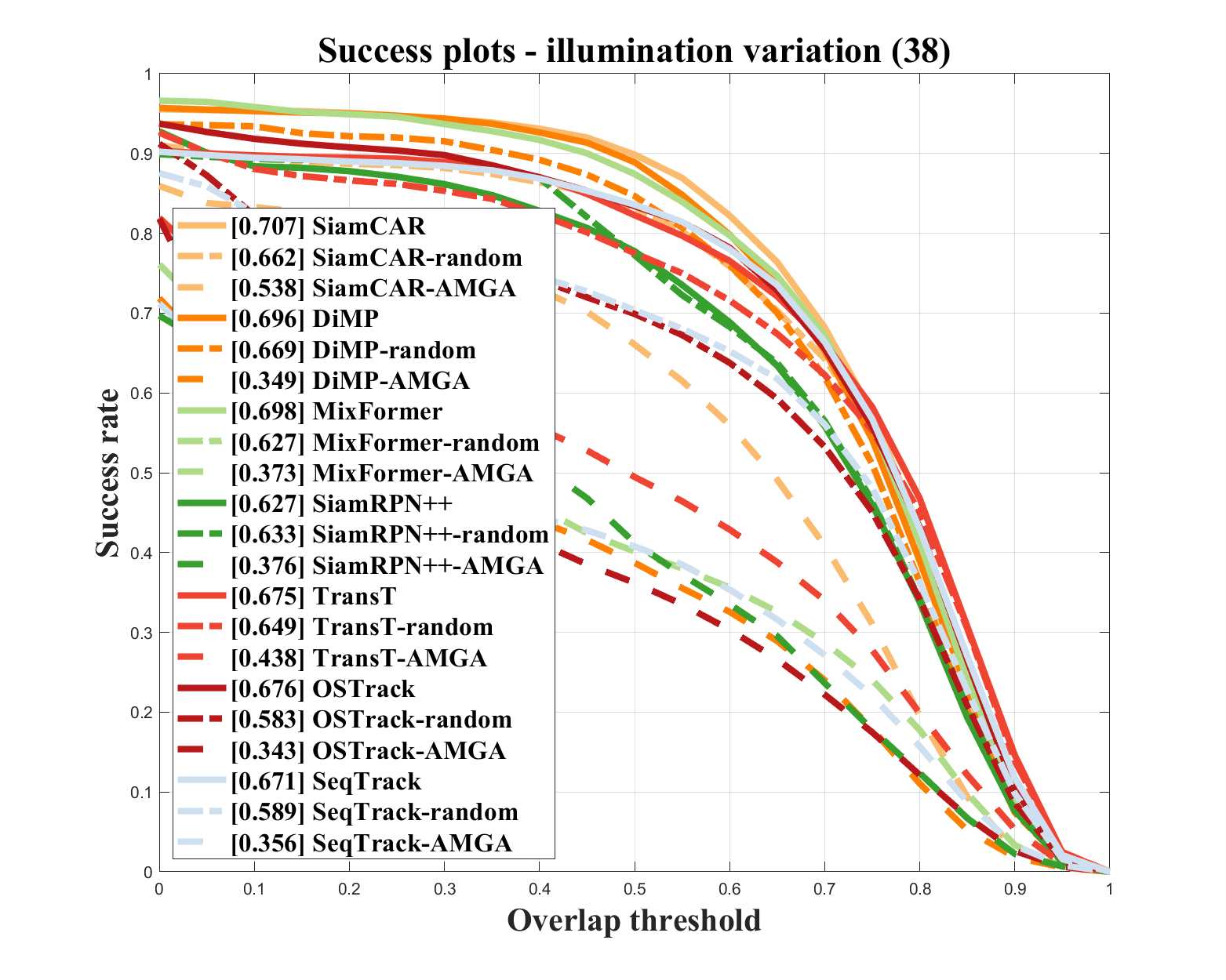}}
		\vfill
		\subfigure{\includegraphics[width=.328\linewidth]{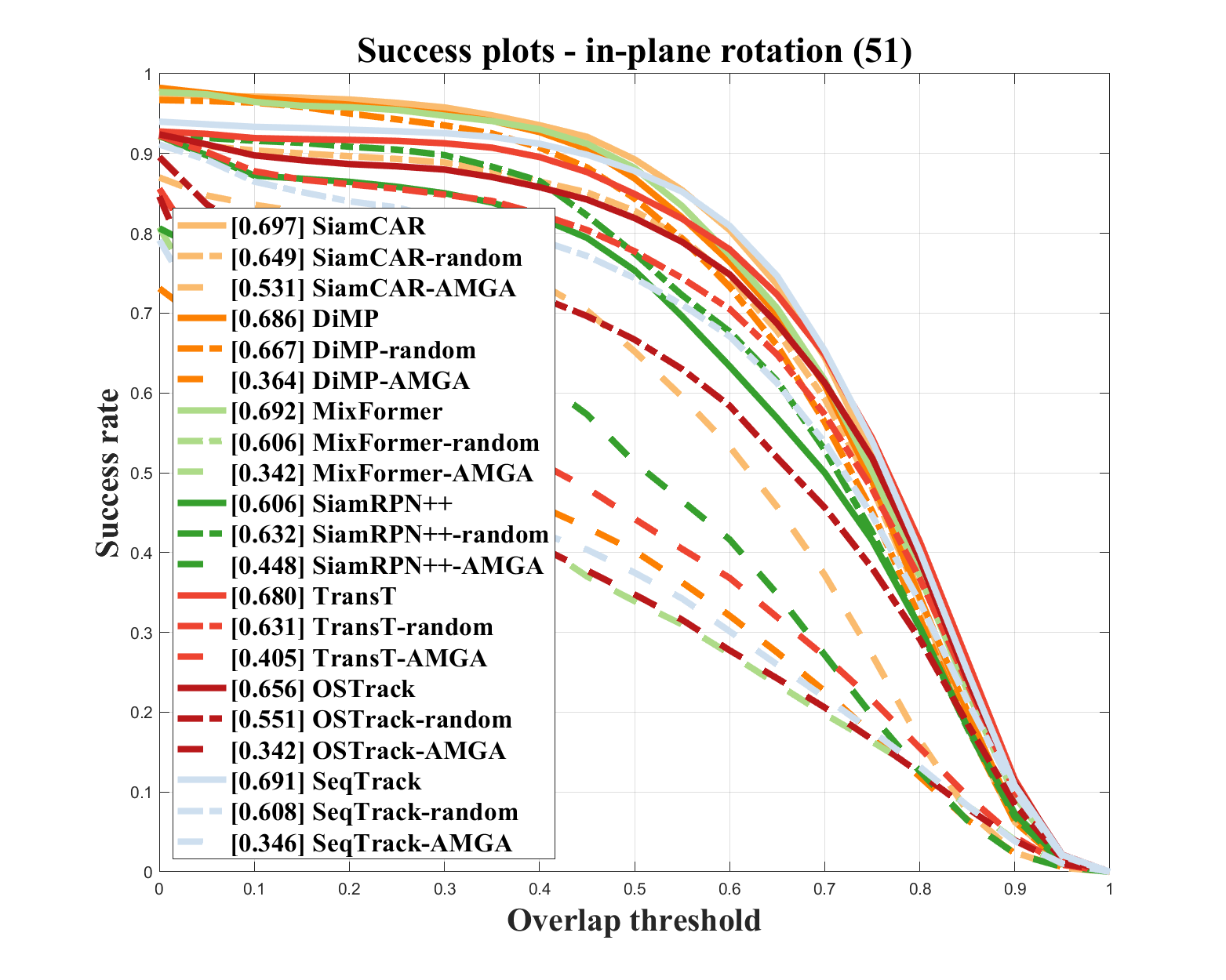}}
		\hspace{0.05em}
		\subfigure{\includegraphics[width=.328\linewidth]{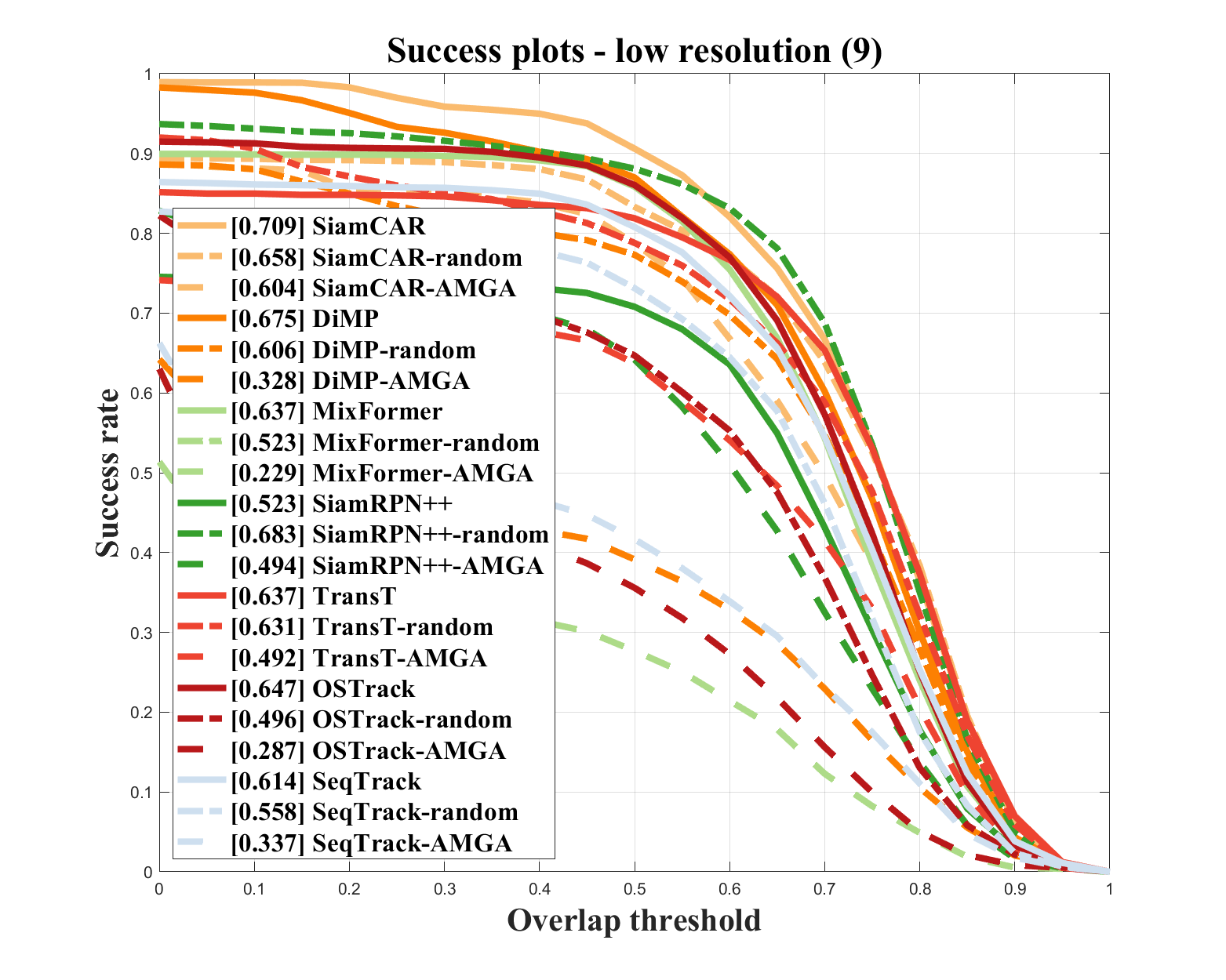}}
		\hspace{0.05em}
		\subfigure{\includegraphics[width=.328\linewidth]{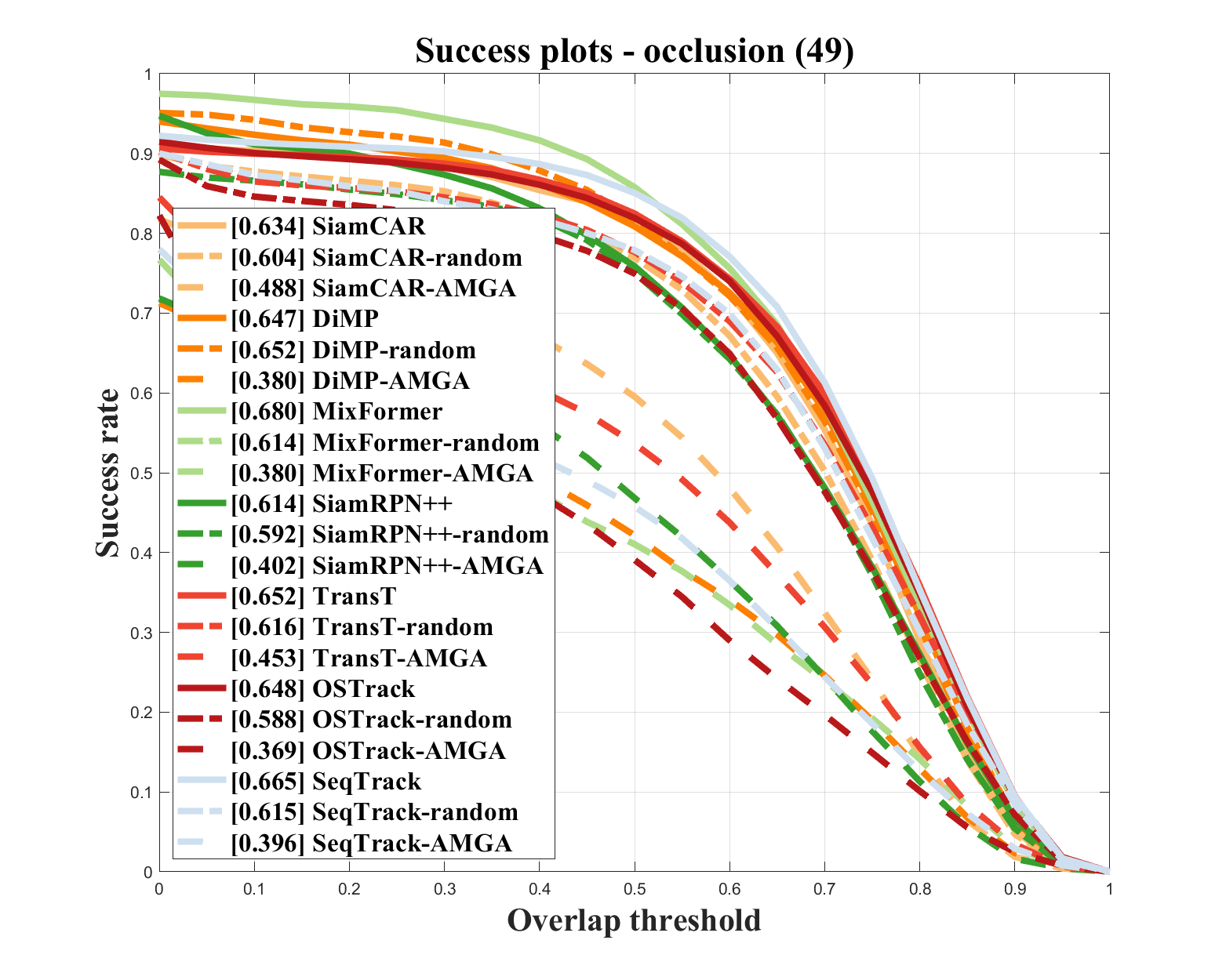}}
		\vfill
		\subfigure{\includegraphics[width=.328\linewidth]{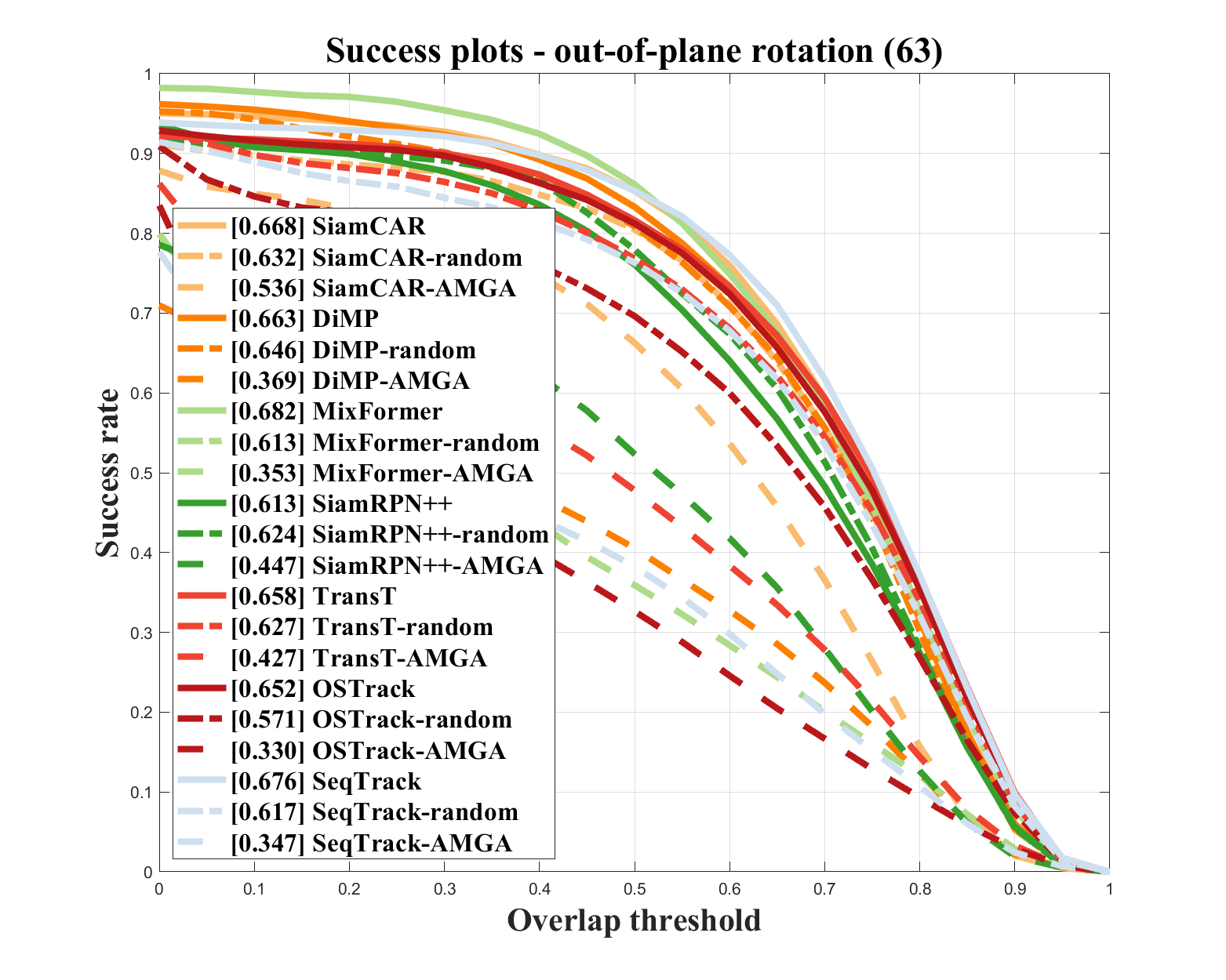}}
		\hspace{0.05em}
		\subfigure{\includegraphics[width=.328\linewidth]{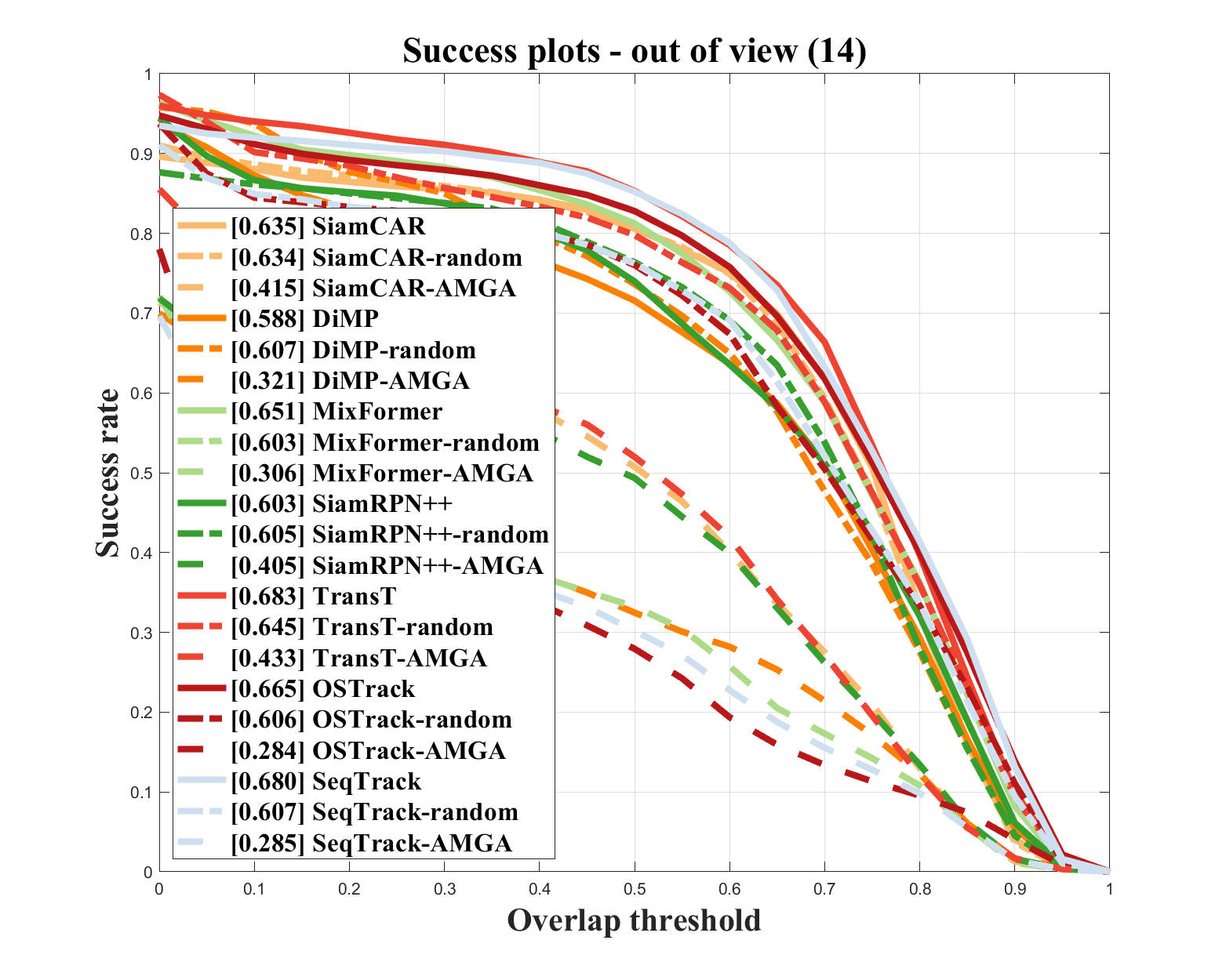}}
		\hspace{0.05em}
		\subfigure{\includegraphics[width=.328\linewidth]{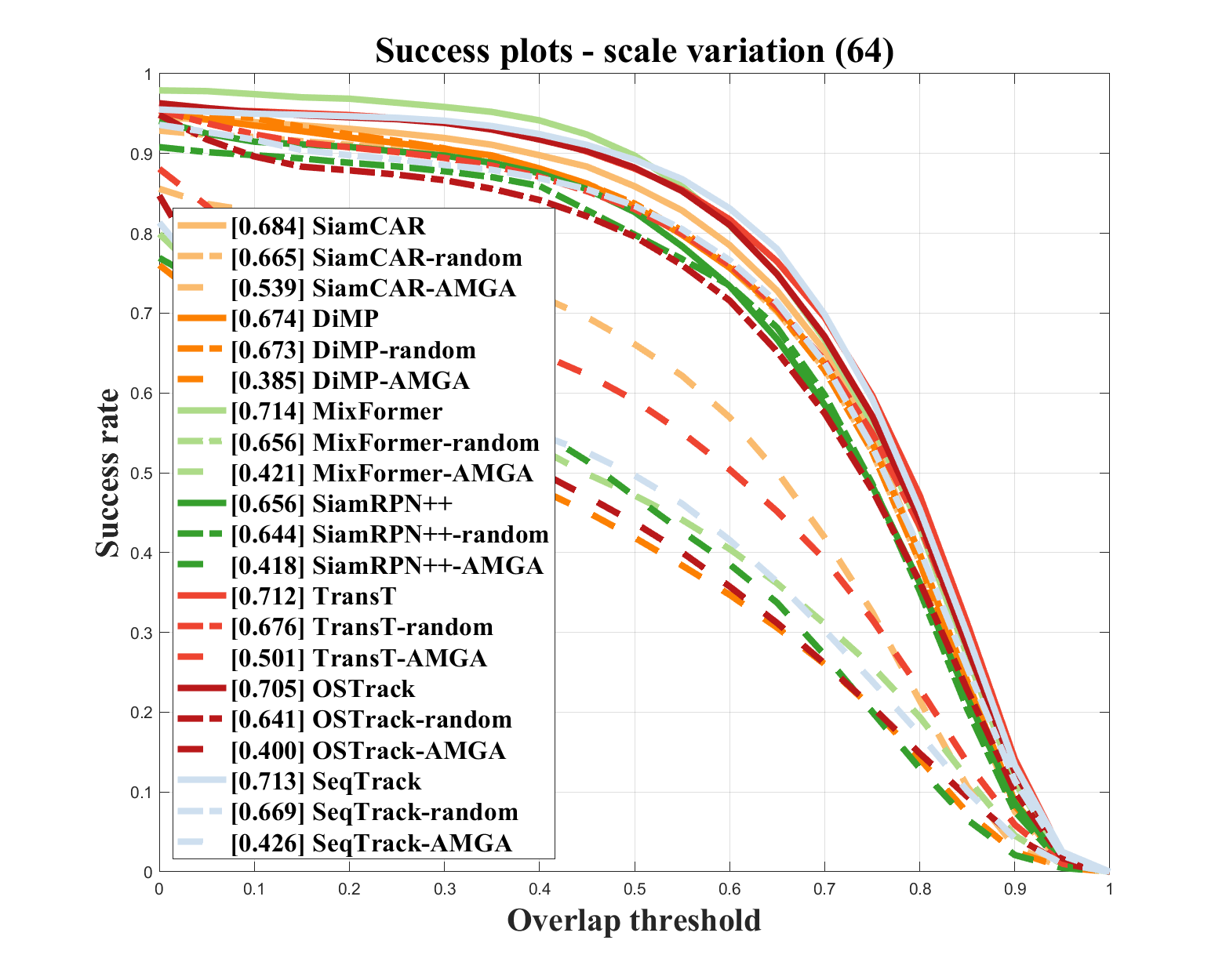}}
	\end{center}
	\vspace{-0.8em}
	\caption{Evaluation on 11 different tracking scenarios of the OTB2015 benchmark dataset\cite{r43}.}
	\label{fig:5}
\end{figure*}

\begin{figure}[t!]
	\begin{center}
		\includegraphics[width=0.6\linewidth]{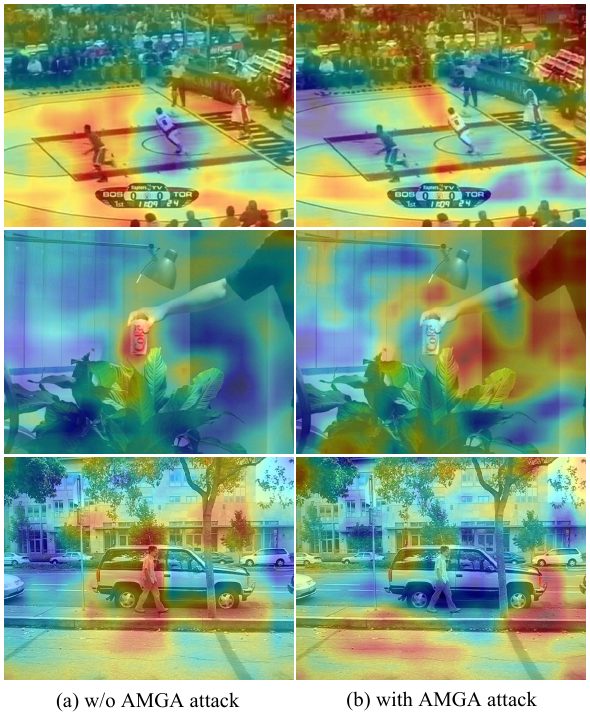}
	\end{center}
	\vspace{-0.8em}
	\caption{Gradient heatmap visualization of SiamRPN++\cite{r9} with and without AMGA attack.}
	\label{fig:4}
\end{figure}

To further investigate the effectiveness of the proposed AMGA method in visual tracking, we designed and conducted a series of ablation studies using SiamRPN++ on the OTB dataset. We performed comparative experiments by selectively removing specific components to evaluate their contributions to attack performance.
We analyzed the independent and combined effects of meta-gradient learning, momentum mechanism, input diversity, and Gaussian smoothing to comprehensively assess their roles in enhancing attack performance. The results are shown in Table \ref{tab:4}. Injecting random noise of equivalent magnitude had minimal impact on SiamRPN++, highlighting that random perturbations are insufficient to significantly disrupt the tracker. Introducing the meta-gradient learning method resulted in a 13.2\% drop in success rate and a 22.3\% drop in precision, demonstrating its ability to effectively interfere with the tracker’s localization capability. Independently applying the input diversity strategy further improved the attack performance. The success rate was reduced to 54.5\%, and precision was decreased by 13.2\%. By using random scaling, padding, and other transformations to input images, input diversity enhanced the robustness of adversarial examples, allowing them to adapt better to various scenarios, such as deformation, occlusion, and scale variation.	When combining the momentum mechanism and Gaussian smoothing with the meta-gradient learning and input diversity, the full AMGA method achieved the most significant reductions in both success rate and precision, demonstrating the synergistic effect of these four strategies in enhancing the overall attack performance of adversarial examples. It is worth noting that when the momentum mechanism was removed from AMGA, the attack performance degraded significantly. This can be attributed to the fact that the momentum mechanism dynamically adjusts the gradient updates, suppressing the gradient oscillations caused by the random model selection, thereby making the generated adversarial examples more stable. Additionally, removing Gaussian smoothing also caused the attack performance to decrease, with a 15.4\% reduction in success rate and a 19.1\% reduction in precision. This indicates that Gaussian smoothing plays an important role in maintaining the imperceptibility of adversarial examples, reducing the high-frequency components of the perturbation, and improving the transferability and stability of the adversarial example across different models and scenarios.

Additionally, we investigated the impact of Gaussian smoothing with different standard deviations on the effectiveness of the AMGA attack. We set the standard deviation ($\sigma$) to 0.5, 1, and 2, and used the peak signal-to-noise ratio (PSNR) and structural similarity index measure (SSIM) as metrics to compare the quality of the adversarial examples generated under different standard deviation settings. In general, the PSNR greater than 30 dB and the SSIM closer to 1 make it difficult for the human vision system to detect differences between the adversarial example and the clean image. The experimental results are shown in Table \ref{tab:6}. Although a minor standard deviation significantly improves the attack effectiveness, it also reduces the imperceptibility of the generated adversarial examples. As the standard deviation increases, the imperceptibility of the adversarial examples improves, but the attack effectiveness slightly weakens. This is because if the standard deviation is too large, it may lead to the loss of some local features, especially detailed textures and edge information. This loss can weaken the adversarial example's ability to interfere with target-specific features, rendering the adversarial example ineffective in complex scenarios, especially for high-resolution images. On the other hand, excessive Gaussian smoothing may cause a large difference in perturbation distribution between consecutive frames, preventing a stable attack from forming across the entire video sequence. Specifically, perturbation with too much smoothing may lead to excessive feature variation between different frames, making the attack ineffective across consecutive frames. This becomes particularly problematic in complex scenarios such as fast move or deformation, where the adversarial example can no longer consistently disrupt the tracker.	Based on the experimental results, we chose a Gaussian smoothing standard deviation of 1 for AMGA, as it balances attack effectiveness and the imperceptibility of the adversarial examples.

\subsection{Comparison with other attack methods}

To further demonstrate the attack effectiveness of AMGA, we conducted comparative experiments on the OTB2015 dataset using SiamRPN++ and compared it with existing white-box and black-box attack methods. The results are shown in Table \ref{tab:5}. Although AMGA does not have direct access to the tracker's architecture and performs slightly below the white-box attack method CSA, it significantly outperforms other black-box attack methods. Specifically, the transfer-based black-box attack method SPARK reduced the success rate and precision of SiamRPN++ by 6.6\% and 2.7\%, respectively, whereas AMGA achieved far more significant reductions in both metrics. Compared with the decision-based black-box attack method IoU, AMGA caused a 21.9\% decrease in success rate, outperforming the 19.6\% decrease of IoU. These results demonstrate that AMGA performs better than existing black-box attack methods, effectively disrupting the tracking performance without knowing the tracker's implementation details.
Although the computational efficiency of AMGA is slightly inferior to other methods due to the higher complexity of meta-learning and multi-model ensemble strategies, this overhead is acceptable considering the significant improvement in attack performance. Overall, AMGA has achieved impressive results in attack effectiveness and efficiency, fully demonstrating its advantages in adversarial attacks for visual tracking.

\subsection{Discussion}

To evaluate the effectiveness of the AMGA method across different tracking scenarios, we conducted scenario-based comparative experiments on the OTB2015 dataset using SiamCAR, SiamRPN, DiMP, MixFormer, TransT, OSTrack, and SeqTrack. Fig.\ref{fig:5} presents the success plots of these trackers under original frames, random noise, and AMGA attacks across 11 different scenarios. AMGA attacks effectively disrupt the feature extraction and matching processes of trackers that rely heavily on local features. By precisely applying adversarial perturbations to the regions of interest, AMGA exploits the dependencies of models on these local features. In complex scenarios, such as occlusions, low resolutions, and scale variations, the appearance of the target object changes significantly, causing models to depend more on the stability and reliability of local features. AMGA leverages this vulnerability, disturbing key local features to degrade the trackers' success rates substantially. However,  the advantage of local perturbation is less pronounced under simpler scenarios, such as illumination variations. This is primarily because illumination variations have a relatively limited impact on feature extraction and similarity matching, as models are typically enhanced for robustness against such variations through data augmentation and regularization during training. These results demonstrate the ability of AMGA to adaptively exploit scenario-specific vulnerabilities in visual trackers, particularly in complex scenarios where local feature stability is critical.

We also visualized the gradient heatmap in Fig.\ref{fig:4}. Without the AMGA attack, the tracker can focus on target-specific regions, such as saliency regions of the target object. However, after the AMGA attack, the gradient heatmap becomes dispersed, and the saliency of target-specific regions decreases. This demonstrates that the AMGA attack successfully disrupts the tracker’s detection and localization capability.

\section{Conclusion}\label{sec:5}

In this paper, we propose AMGA, a meta-learning-based adversarial attack method combined with a multi-model ensemble strategy for generating adversarial examples in visual tracking. Without relying on the network architecture information of the tracker, we iteratively optimize adversarial perturbations based on the classification results of multiple pre-trained deep learning models applied to the input frame. This method effectively generates adversarial examples with strong transferability and robustness, successfully addressing the challenges of cross-model transfer and stability in attack performance. By integrating momentum mechanism, Gaussian smoothing, and input diversity, AMGA adaptively adjusts the updating direction of the adversarial perturbations, ensuring that they effectively transfer to subsequent frames. This significantly enhances the interference capability against the tracker. AMGA is particularly suited for widely used deep learning-based trackers, as it can generate adversarial examples with strong generalization abilities without requiring access to internal gradient information. This extends the applicability of adversarial attacks in visual tracking. To validate the generality and effectiveness of the proposed AMGA method, we applied it to seven state-of-the-art trackers, which cover different network architectures and tracking strategies. Extensive experiments on multiple benchmark datasets demonstrate that AMGA exhibits superior attack performance across various trackers. Nevertheless, the transferability and robustness of adversarial examples can be further enhanced.
Although AMGA significantly improves the transferability of the generated adversarial examples through a combination of various strategies, it isn't easy to exhaustively cover all deep learning models used in the visual tracking domain. Its performance may be limited by the diversity of the model repository and the architectural differences of the trackers, which could result in poorer performance on some specific tasks. As for existing adversarial defense methods, such as model pruning, adversarial detection, and input transformations, while they can mitigate the attack effects of AMGA in some scenarios, AMGA still maintains high attack performance in a broader range of scenarios due to the incorporation of strategies like multi-model ensemble, momentum mechanism, and Gaussian smoothing in the generation of adversarial examples. In our future research, while further improving the attack performance of AMGA, we will also focus on how to counter existing and emerging adversarial defense methods to enhance the attack robustness of AMGA.

\bibliographystyle{num}
\bibliography{bibliography}

\begin{thebibliography}{10}
\expandafter\ifx\csname url\endcsname\relax
  \def\url#1{\texttt{#1}}\fi
\expandafter\ifx\csname urlprefix\endcsname\relax\def\urlprefix{URL }\fi
\expandafter\ifx\csname href\endcsname\relax
  \def\href#1#2{#2} \def\path#1{#1}\fi

\bibitem{r1}
F.~Chen, X.~Wang, Y.~Zhao, S.~Lv, X.~Niu, Visual object tracking: A survey,
  Computer Vision and Image Understanding 222 (2022) 103508.

\bibitem{r3}
V.~Jadeja, A.~Rao, A.~Srivastava, S.~Singh, P.~Chaturvedi, G.~Bhardwaj,
  Convolutional neural networks: A comprehensive review of architectures and
  application, in: 2023 6th International Conference on Contemporary Computing
  and Informatics (IC3I), Vol.~6, IEEE, 2023, pp. 460--467.

\bibitem{r4}
W.~Rawat, Z.~Wang, Deep convolutional neural networks for image classification:
  A comprehensive review, Neural computation 29~(9) (2017) 2352--2449.

\bibitem{rar}
P.~Gao, Q.~Zhang, F.~Wang, L.~Xiao, H.~Fujita, Y.~Zhang, Learning reinforced
  attentional representation for end-to-end visual tracking, Information
  Sciences 517 (2020) 52--67.

\bibitem{r8}
L.~Bertinetto, J.~Valmadre, J.~F. Henriques, A.~Vedaldi, P.~H. Torr,
  Fully-convolutional siamese networks for object tracking, in: Computer
  Vision--ECCV 2016 Workshops: Amsterdam, The Netherlands, October 8-10 and
  15-16, 2016, Proceedings, Part II 14, Springer, 2016, pp. 850--865.

\bibitem{satin}
P.~Gao, R.~Yuan, F.~Wang, L.~Xiao, H.~Fujita, Y.~Zhang, Siamese attentional
  keypoint network for high performance visual tracking, Knowledge-based
  systems 193 (2020) 105448.

\bibitem{r6}
B.~Li, J.~Yan, W.~Wu, Z.~Zhu, X.~Hu, High performance visual tracking with
  siamese region proposal network, in: Proceedings of the IEEE/CVF conference
  on computer vision and pattern recognition, 2018, pp. 8971--8980.

\bibitem{siamextr}
P.~Gao, X.-Y. Zhang, X.-L. Yang, F.~Gao, H.~Fujita, F.~Wang, Robust visual
  tracking with extreme point graph-guided annotation: Approach and experiment,
  Expert Systems with Applications 238 (2024) 122013.

\bibitem{r2}
I.~J. Goodfellow, J.~Shlens, C.~Szegedy, Explaining and harnessing adversarial
  examples, arXiv preprint arXiv:1412.6572.

\bibitem{r11}
A.~Kurakin, I.~J. Goodfellow, S.~Bengio, Adversarial examples in the physical
  world, in: Artificial intelligence safety and security, Chapman and Hall/CRC,
  2018, pp. 99--112.

\bibitem{r40}
L.~Zhang, Y.~Zhou, Y.~Yang, X.~Gao, Meta invariance defense towards
  generalizable robustness to unknown adversarial attacks, IEEE Transactions on
  Pattern Analysis and Machine Intelligence.

\bibitem{r21}
T.~Hospedales, A.~Antoniou, P.~Micaelli, A.~Storkey, Meta-learning in neural
  networks: A survey, IEEE transactions on pattern analysis and machine
  intelligence 44~(9) (2021) 5149--5169.

\bibitem{r13}
C.~Szegedy, V.~Vanhoucke, S.~Ioffe, J.~Shlens, Z.~Wojna, Rethinking the
  inception architecture for computer vision, in: Proceedings of the IEEE/CVF
  conference on computer vision and pattern recognition, 2016, pp. 2818--2826.

\bibitem{r14}
K.~He, X.~Zhang, S.~Ren, J.~Sun, Deep residual learning for image recognition,
  in: Proceedings of the IEEE/CVF conference on computer vision and pattern
  recognition, 2016, pp. 770--778.

\bibitem{r15}
M.~Sandler, A.~Howard, M.~Zhu, A.~Zhmoginov, L.-C. Chen, Mobilenetv2: Inverted
  residuals and linear bottlenecks, in: Proceedings of the IEEE/CVF conference
  on computer vision and pattern recognition, 2018, pp. 4510--4520.

\bibitem{r16}
G.~Huang, Z.~Liu, L.~Van Der~Maaten, K.~Q. Weinberger, Densely connected
  convolutional networks, in: Proceedings of the IEEE/CVF conference on
  computer vision and pattern recognition, 2017, pp. 4700--4708.

\bibitem{r17}
K.~Simonyan, Very deep convolutional networks for large-scale image
  recognition, arXiv preprint arXiv:1409.1556.

\bibitem{r18}
A.~Krizhevsky, I.~Sutskever, G.~E. Hinton, Imagenet classification with deep
  convolutional neural networks, Communications of the ACM 60~(6) (2017)
  84--90.

\bibitem{r19}
F.~N. Iandola, S.~Han, M.~W. Moskewicz, K.~Ashraf, W.~J. Dally, K.~Keutzer,
  Squeezenet: Alexnet-level accuracy with 50x fewer parameters and $<$0.5mb
  model size, arXiv preprint arXiv:1602.07360.

\bibitem{r20}
S.~Zagoruyko, Wide residual networks, arXiv preprint arXiv:1605.07146.

\bibitem{r10}
D.~Guo, J.~Wang, Y.~Cui, Z.~Wang, S.~Chen, Siamcar: Siamese fully convolutional
  classification and regression for visual tracking, in: Proceedings of the
  IEEE/CVF conference on computer vision and pattern recognition, 2020, pp.
  6269--6277.

\bibitem{r9}
B.~Li, W.~Wu, Q.~Wang, F.~Zhang, J.~Xing, J.~Yan, Siamrpn++: Evolution of
  siamese visual tracking with very deep networks, in: Proceedings of the
  IEEE/CVF conference on computer vision and pattern recognition, 2019, pp.
  4282--4291.

\bibitem{r22}
G.~Bhat, M.~Danelljan, L.~V. Gool, R.~Timofte, Learning discriminative model
  prediction for tracking, in: Proceedings of the IEEE/CVF international
  conference on computer vision, 2019, pp. 6182--6191.

\bibitem{r23}
Y.~Cui, C.~Jiang, L.~Wang, G.~Wu, Mixformer: End-to-end tracking with iterative
  mixed attention, in: Proceedings of the IEEE/CVF conference on computer
  vision and pattern recognition, 2022, pp. 13608--13618.

\bibitem{r24}
X.~Chen, B.~Yan, J.~Zhu, D.~Wang, X.~Yang, H.~Lu, Transformer tracking, in:
  Proceedings of the IEEE/CVF conference on computer vision and pattern
  recognition, 2021, pp. 8126--8135.

\bibitem{r48}
B.~Ye, H.~Chang, B.~Ma, S.~Shan, X.~Chen, Joint feature learning and relation
  modeling for tracking: A one-stream framework, in: European conference on
  computer vision, Springer, 2022, pp. 341--357.

\bibitem{r49}
X.~Chen, H.~Peng, D.~Wang, H.~Lu, H.~Hu, Seqtrack: Sequence to sequence
  learning for visual object tracking, in: Proceedings of the IEEE/CVF
  conference on computer vision and pattern recognition, 2023, pp.
  14572--14581.

\bibitem{r43}
Y.~Wu, J.~Lim, M.-H. Yang, Object tracking benchmark, IEEE Transactions on
  Pattern Analysis and Machine Intelligence 37~(09) (2015) 1834--1848.

\bibitem{r45}
L.~Huang, X.~Zhao, K.~Huang, Got-10k: A large high-diversity benchmark for
  generic object tracking in the wild, IEEE Transactions on Pattern Analysis
  and Machine Intelligence 43~(5) (2019) 1562--1577.

\bibitem{r44}
H.~Fan, H.~Bai, L.~Lin, F.~Yang, P.~Chu, G.~Deng, S.~Yu, M.~Huang, J.~Liu,
  Y.~Xu, et~al., Lasot: A high-quality large-scale single object tracking
  benchmark, International Journal of Computer Vision 129~(2) (2021) 439--461.

\bibitem{r27}
A.~Vaswani, Attention is all you need, Advances in Neural Information
  Processing Systems.

\bibitem{r25}
L.~Zhang, A.~Gonzalez-Garcia, J.~V.~D. Weijer, M.~Danelljan, F.~S. Khan,
  Learning the model update for siamese trackers, in: Proceedings of the
  IEEE/CVF international conference on computer vision, 2019, pp. 4010--4019.

\bibitem{r26}
M.~Danelljan, L.~V. Gool, R.~Timofte, Probabilistic regression for visual
  tracking, in: Proceedings of the IEEE/CVF conference on computer vision and
  pattern recognition, 2020, pp. 7183--7192.

\bibitem{r28}
F.~Tramer, A.~Kurakin, N.~Papernot, I.~Goodfellow, D.~Boneh, P.~McDaniel,
  Ensemble adversarial training: Attacks and defenses, arXiv preprint
  arXiv:1705.07204.

\bibitem{r30}
Y.~Dong, F.~Liao, T.~Pang, X.~Hu, J.~Zhu, Discovering adversarial examples with
  momentum, arXiv preprint arXiv:1710.06081 5.

\bibitem{r31}
K.~Ren, T.~Zheng, Z.~Qin, X.~Liu, Adversarial attacks and defenses in deep
  learning, Engineering 6~(3) (2020) 346--360.

\bibitem{r32}
M.~Lei, H.~Song, J.~Fan, D.~Xiao, D.~Ai, Y.~Gu, J.~Yang, Gaa: Ghost adversarial
  attack for object tracking, IEEE Transactions on Emerging Topics in
  Computational Intelligence.

\bibitem{r33}
S.~Jia, C.~Ma, Y.~Song, X.~Yang, M.-H. Yang, Robust deep object tracking
  against adversarial attacks, International Journal of Computer Vision (2024)
  1--20.

\bibitem{r34}
S.~Liang, X.~Wei, S.~Yao, X.~Cao, Efficient adversarial attacks for visual
  object tracking, in: Computer Vision--ECCV 2020: 16th European Conference,
  Glasgow, UK, August 23--28, 2020, Proceedings, Part XXVI 16, Springer, 2020,
  pp. 34--50.

\bibitem{r35}
S.~Zhao, T.~Xu, X.-J. Wu, J.~Kittler, Pluggable attack for visual object
  tracking, IEEE Transactions on Information Forensics and Security 19 (2023)
  1227--1240.

\bibitem{r36}
X.~Chen, X.~Yan, F.~Zheng, Y.~Jiang, S.-T. Xia, Y.~Zhao, R.~Ji, One-shot
  adversarial attacks on visual tracking with dual attention, in: Proceedings
  of the IEEE/CVF conference on computer vision and pattern recognition, 2020,
  pp. 10176--10185.

\bibitem{r37}
Q.~Guo, Z.~Cheng, F.~Juefei-Xu, L.~Ma, X.~Xie, Y.~Liu, J.~Zhao, Learning to
  adversarially blur visual object tracking, in: Proceedings of the IEEE/CVF
  international conference on computer vision, 2021, pp. 10839--10848.

\bibitem{r39}
Z.~Yuan, J.~Zhang, Y.~Jia, C.~Tan, T.~Xue, S.~Shan, Meta gradient adversarial
  attack, in: Proceedings of the IEEE/CVF International Conference on Computer
  Vision, 2021, pp. 7748--7757.

\bibitem{r46}
B.~Yan, D.~Wang, H.~Lu, X.~Yang, Cooling-shrinking attack: Blinding the tracker
  with imperceptible noises, in: Proceedings of the IEEE/CVF conference on
  computer vision and pattern recognition, 2020, pp. 990--999.

\bibitem{r47}
Q.~Guo, X.~Xie, F.~Juefei-Xu, L.~Ma, Z.~Li, W.~Xue, W.~Feng, Y.~Liu, Spark:
  Spatial-aware online incremental attack against visual tracking, in: European
  conference on computer vision, Springer, 2020, pp. 202--219.

\bibitem{r12}
S.~Jia, Y.~Song, C.~Ma, X.~Yang, Iou attack: Towards temporally coherent
  black-box adversarial attack for visual object tracking, in: Proceedings of
  the IEEE/CVF conference on computer vision and pattern recognition, 2021, pp.
  6709--6718.

\end{thebibliography}


\end{document}